\documentclass[conference]{IEEEtran}
\IEEEoverridecommandlockouts
\usepackage{cite}
\usepackage{amsmath,amssymb,amsfonts,amsthm}
\usepackage{algorithmic}
\usepackage{graphicx}
\usepackage{textcomp}
\usepackage[dvipsnames]{xcolor}
\usepackage{etoolbox}
\usepackage{multirow}
\usepackage{array}
\usepackage{fancybox}
\usepackage{adjustbox}
\usepackage{hyperref}
\usepackage[nameinlink,noabbrev,capitalize]{cleveref}
\usepackage{titlesec}
\usepackage{booktabs}
\usepackage{fancybox}
\usepackage{subcaption}
\usepackage{empheq}
\usepackage{enumitem}
\usepackage{ifthen}
\usepackage{verbatim}
\usepackage{algorithm}
\usepackage{xspace}
\usepackage{setspace}
\usepackage{stmaryrd}

\usepackage{pifont}
\usepackage{verbatim,ifthen,alltt}
\usepackage{xfrac}
\usepackage{tikz}
\usepackage{pgf}
\usepackage{forest}

\usetikzlibrary{fit,calc,tikzmark}
\usetikzlibrary{arrows,decorations.pathmorphing,decorations.pathreplacing,decorations.footprints,fadings,calc,trees,mindmap,shadows,decorations.text,patterns,positioning,shapes}
\usetikzlibrary{arrows,shadows,backgrounds}
\usetikzlibrary{arrows.meta}
\usetikzlibrary{positioning,fit,automata,matrix}
\usetikzlibrary{shapes.symbols,shapes.misc,shapes.arrows}

\def\BibTeX{{\rm B\kern-.05em{\sc i\kern-.025em b}\kern-.08em
    T\kern-.1667em\lower.7ex\hbox{E}\kern-.125emX}}

\definecolor{gray}{rgb}{.4,.4,.4}
\definecolor{midgrey}{rgb}{0.5,0.5,0.5}
\definecolor{middarkgrey}{rgb}{0.35,0.35,0.35}
\definecolor{darkgrey}{rgb}{0.3,0.3,0.3}
\definecolor{darkred}{rgb}{0.7,0.1,0.1}
\definecolor{midblue}{rgb}{0.2,0.2,0.7}
\definecolor{darkblue}{rgb}{0.1,0.1,0.5}
\definecolor{defseagreen}{cmyk}{0.69,0,0.50,0}

\definecolor{tyellow1}{HTML}{FCE94F}
\definecolor{tyellow2}{HTML}{EDD400}
\definecolor{tyellow3}{HTML}{C4A000}
\definecolor{torange1}{HTML}{FCAF3E}
\definecolor{torange2}{HTML}{F57900}
\definecolor{torange3}{HTML}{C35C00}
\definecolor{tbrown1}{HTML}{E9B96E}
\definecolor{tbrown2}{HTML}{C17D11}
\definecolor{tbrown3}{HTML}{8F5902}
\definecolor{tgreen1}{HTML}{8AE234}
\definecolor{tgreen2}{HTML}{73D216}
\definecolor{tgreen3}{HTML}{4E9A06}
\definecolor{tblue1}{HTML}{729FCF}
\definecolor{tblue2}{HTML}{3465A4}
\definecolor{tblue3}{HTML}{204A87}
\definecolor{tpurple1}{HTML}{AD7FA8}
\definecolor{tpurple2}{HTML}{75507B}
\definecolor{tpurple3}{HTML}{5C3566}
\definecolor{tred1}{HTML}{EF2929}
\definecolor{tred2}{HTML}{CC0000}
\definecolor{tred3}{HTML}{A40000}
\definecolor{tlgray1}{HTML}{EEEEEC}
\definecolor{tlgray2}{HTML}{D3D7CF}
\definecolor{tlgray3}{HTML}{BABDB6}
\definecolor{tdgray1}{HTML}{888A85}
\definecolor{tdgray2}{HTML}{555753}
\definecolor{tdgray3}{HTML}{2E3436}

\newcommand{\hlight}[1]{{\color{darkred}#1}}
\newcommand{\rhlight}[1]{\hlight{#1}}
\newcommand{\dghlight}[1]{{\color[RGB]{0,120,0}#1}}

\newboolean{nonanon}               
\setboolean{nonanon}{true}         %
\newboolean{mkcompact}             
\setboolean{mkcompact}{true}       %
\newboolean{squeezeit}             
\setboolean{squeezeit}{true}       %
\crefname{prop}{Proposition}{Propositions}
\Crefname{prop}{Proposition}{Propositions}
\crefname{defn}{Definition}{Definitions}
\Crefname{defn}{Definition}{Definitions}
\crefname{cor}{Corollary}{Corollaries}
\Crefname{cor}{Corollary}{Corollaries}
\crefname{exmpl}{Example}{Examples}
\Crefname{exmpl}{Example}{Examples}
\Crefname{theorem}{Theorem}{Theorem}
\crefname{enumi}{}{}

\newtheoremstyle{nutstyle}
{3.0pt} 
{3.0pt} 
{} 
{} 
{\bfseries} 
{.} 
{.5em} 
{} 

\theoremstyle{nutstyle}
\newtheorem{prop}{Proposition}
\newtheorem{defn}{Definition}
\newtheorem{cor}{Corollary}

\newtheoremstyle{nuxstyle}
{3.0pt} 
{3.0pt} 
{} 
{} 
{\itshape} 
{.} 
{.5em} 
{} 

\theoremstyle{nuxstyle}

\newtheorem{example}{Example}

\newcommand{\todoF}[2]{}

\newcommand{\fml}[1]{{\mathcal{#1}}}

\newcommand{\tn}[1]{\textnormal{#1}}
\newcommand{\mbf}[1]{\ensuremath\mathbf{#1}}
\newcommand{\msf}[1]{\ensuremath\mathsf{#1}}
\newcommand{\mbb}[1]{\ensuremath\mathbb{#1}}

\newcommand{\tbf}[1]{\textbf{#1}}

\newcommand{\True}{\textbf{true}}
\newcommand{\False}{\textbf{false}}

\DeclareMathOperator*{\chkco}{\textnormal{\bfseries{\textsf{CO}}}\!}
\newcommand{\consistent}[1]{\chkco\left(#1\right)}

\newcommand{\relevant}{\oper{Relevant}}
\newcommand{\irrelevant}{\oper{Irrelevant}}

\newcommand{\sv}{\ensuremath\msf{Sv}}

\definecolor{darkslategray}{rgb}{0.18, 0.31, 0.31} 
\definecolor{platinum}{rgb}{0.9, 0.89, 0.89} 
\definecolor{gray}{rgb}{.4,.4,.4}
\definecolor{midgrey}{rgb}{0.5,0.5,0.5}
\definecolor{middarkgrey}{rgb}{0.35,0.35,0.35}
\definecolor{darkgrey}{rgb}{0.3,0.3,0.3}
\definecolor{darkred}{rgb}{0.7,0.1,0.1}
\definecolor{midblue}{rgb}{0.2,0.2,0.7}
\definecolor{darkblue}{rgb}{0.1,0.1,0.5}
\definecolor{defseagreen}{cmyk}{0.69,0,0.50,0}
\newcommand{\jnote}[1]{\medskip\noindent$\llbracket$\textcolor{darkred}{joao}: \emph{\textcolor{middarkgrey}{#1}}$\rrbracket$\medskip}

\newcommand{\jnoteF}[1]{}

\newcommand{\oper}[1]{\ensuremath\textnormal{\small{\textsf{#1}}}}

\newcounter{tableeqn}[table]

\DeclareMathOperator*{\limply}{\rightarrow}

\newcommand{\bigland}{\ensuremath\bigwedge}
\newcommand{\waxp}{\ensuremath\mathsf{WAXp}}
\newcommand{\wcxp}{\ensuremath\mathsf{WCXp}}
\newcommand{\axp}{\ensuremath\mathsf{AXp}}
\newcommand{\cxp}{\ensuremath\mathsf{CXp}}

\newcommand{\mailtodomain}[1]{\href{mailto:#1@ciencias.ulisboa.pt}{\texttt{\nolinkurl{#1}}}}

\titleformat{\paragraph}[runin]
{\normalfont\bfseries}{}{0pt}{}

\newcolumntype{L}[1]{>{\raggedright\let\newline\\\arraybackslash\hspace{0pt}}m{#1}}
\newcolumntype{C}[1]{>{\centering\let\newline\\\arraybackslash\hspace{0pt}}m{#1}}
\newcolumntype{R}[1]{>{\raggedleft\let\newline\\\arraybackslash\hspace{0pt}}m{#1}}

\hypersetup{
  colorlinks = true,
  linkcolor = MidnightBlue,
  urlcolor  = MidnightBlue,
  citecolor = MidnightBlue,
  anchorcolor = MidnightBlue
}

\titleclass{\acks}{straight}[\subsection]

\titleformat{\acks}[runin]
  {\normalfont\bfseries}{}{0em}{Acknowledgments.}
\titlespacing*{\acks}{0pt}{3.25ex plus 1ex minus .2ex}{1.5ex plus .2ex}

\tikzset{
  0 my edge/.style={densely dashed, my edge},
  my edge/.style={-{Stealth[]}},
}

\begin{document}

%
\title{Disproving XAI Myths with Formal Methods --\\
  Initial Results
  \thanks{
    This work was supported by the AI Interdisciplinary Institute
    ANITI, funded by the French program ``Investing for the Future --
    PIA3'' under Grant agreement no.\ ANR-19-PI3A-0004,
    and
    by the H2020-ICT38 project COALA ``Cognitive Assisted agile
    manufacturing for a Labor force supported by trustworthy Artificial
    intelligence''.}
}


\author{\IEEEauthorblockN{Joao Marques-Silva}
\IEEEauthorblockA{\textit{IRIT, CNRS} \\
Toulouse, France \\
joao.marques-silva@irit.fr}
}

\maketitle

\maketitle

\begin{abstract}
  The advances in Machine Learning (ML) in recent years have been both
  impressive and far-reaching. However, the deployment of ML models is
  still impaired by a lack of trust in how the best-performing ML models
  make predictions. The issue of lack of trust is even more acute in
  the uses of ML models in high-risk or safety-critical domains.
  eXplainable artificial intelligence (XAI) is at the core of ongoing
  efforts for delivering trustworthy AI. Unfortunately, XAI is riddled
  with critical misconceptions, that foster distrust instead of
  building trust. This paper details some of the most visible
  misconceptions in XAI, and shows how formal methods have been used,
  both to disprove those misconceptions, but also to devise
  practically effective alternatives.
  %
\end{abstract}


\section{Introduction} \label{sec:intro}

The advances in Machine Learning (ML) in recent years have been
absolutely remarkable~\cite{bengio-nature15,bengio-cacm21}, and are
posed to impact many aspects of society.
However, the most promising ML models exhibit important shortcomings.
First, ML models can be brittle, in that small changes to the inputs 
can cause unexpected changes on the output~\cite{goodfellow-iclr14}.
The assessment of \emph{robustness} of ML models aims to confirm that
ML models are not brittle, and has been an active area of research in
recent years.
Second, ML models can exhibit bias, and well-known examples exist of
biased decision making~\cite{propublica16,theverge20}. Fairness of ML
models is another active area of
research~\cite{galstyan-acmcs21,shmueli-acmcs22,fair-bk22}, with the
use of formal methods also proposed~\cite{icshms-cp20}.
Third, ML models are inscrutable in their operation, and as a result
are often referred to as black-box ML models.
%
%
Both shortcomings of ML models motivate distrust by human decision
makers, and so represent critical limitations to the widespread 
deployment of ML models in many practical applications, especially
among those that are deemed high-risk or safety-critical.
In recent years, there have been calls for the verification of
AI~\cite{seshia-cacm22}, aiming to target both the lack of robustness
of ML models and their inscrutability.


eXplainable Artificial Intelligence
(XAI)~\cite{gunning-tr16,gunning-aimag19,gunning-sr19}, addresses the
shortcoming of inscrutability, in that it broadly aims to help human
decision makers in understanding the operation of ML models. 
The perceived importance of XAI motivated a large body of research,
covering a wide range of topics, with the overarching goal of helping
to build trust in the use of ML models.
However, most work on XAI is not supported by formal guarantees of
rigor. We will refer to such XAI efforts as \emph{non-formal XAI}.
The most visible exception to this state of affairs is formal XAI,
which builds on rigorously defined notions of explainability, and the
use of automated reasoning for computing explanations. Formal XAI is
not without limitations, with the best-known being scalability with
respect to complex ML models.

This paper makes a case for the importance of formal XAI, but showing
how formal reasoning about explanations has served to refute three of
the most established myths of non-formal XAI.

\paragraph{Myth \#01 -- intrinsic interpretability.}~
%
Some ML models are widely accepted to be
interpretable~\cite{molnar-bk20}, even if there is no rigorous
definition of what \emph{interpretable} means~\cite{lipton-cacm18}.
Concrete examples of so-called interpretable models include decision
trees, lists and sets.
The \emph{interpretability} of these ML models has motivated recent
works to advocate for their use in high-risk
applications~\cite{rudin-naturemi19,rudin-ss22}.

In recent
work~\cite{iims-corr20,hiims-kr21,iims-jair22,msi-fai23}, we 
demonstrated that these so-called interpretable models will not
provide interpretable explanations as long as these relate with
explanation succinctness. \cref{sec:myth01} overviews the main results
reported in recent work, which disprove existing claims about
interpretability of some ML models.

\paragraph{Myth \#02 -- model-agnostic explanations.}~
%
The most popular approaches for explaining ML models using
\emph{feature selection} (i.e.\ sets of features) are referred to as
\emph{model-agnostic}. Essentially, a model-agnostic explainability
approach ignores the ML model, and proposes an explanation given the
training data, and an estimated distribution of the input. One of the
best-known approaches for model-agnostic explainability reports
explanations as sets of features which explain the prediction.
Anchors is arguable one of the best known model-agnostic
explainers~\cite{guestrin-aaai18}.

In recent work~\cite{inms-corr19,ignatiev-ijcai20,yisnms-aaai23},
we demonstrated that the sets of features reported by
model-agnostic approaches are in the vast majority of cases
\emph{not} sufficient for the prediction, and so can mislead human
decision makers by overlooking critical features.
\cref{sec:myth02} overviews the main results reported in recent work,
which disprove the ability of model-agnostic methods for computing
sufficient reasons for explanations.

\paragraph{Myth \#03 -- explainability with Shapley values.}~
%
The best-known alternative to feature selection is \emph{feature
attribution}, where each feature receives a score of its relative
importance for a prediction. Methods of feature attribution rank among
the best-known and more widely used explainability
approaches~\cite{muller-plosone15,guestrin-kdd16a,lundberg-nips17,xai-bk19}.
Among these, SHAP~\cite{lundberg-nips17} builds on the use of Shapley
values~\cite{shapley-ctg53}. The success of Shapley values for
explainability is illustrated by a number of works. Furthermore,
Shapley values for explainability provide a theoretical justification
for SHAP-like feature attribution methods.

In recent work~\cite{hms-corr23}, we demonstrated that Shapley
values for explainability, as proposed by a number of
authors~\cite{kononenko-jmlr10,kononenko-kis14,lundberg-nips17,lundberg-naturemi20},
will provide misleading information about feature importance.
To derive this result, we characterized features in terms of being
relevant or irrelevant for rigorous explanations based on feature
selection. We then showed that irrelevant features for explanations
based on feature selection, i.e.\ features that do not occur in
\emph{any} (irreducible) explanation based on selection of features,
will exhibit an absolute Shapley value larger than relevant features
for explanations based on feature selection.
\cref{sec:myth03} overviews the main results reported in recent work,
which disprove the adequacy of Shapley values for explainability.

\paragraph{Organization.}~
%
The paper is organized as follows.
\cref{sec:prelim} introduces the notation and definitions used
throughout.
\cref{sec:xai} provides a brief account of XAI, emphasizing formal
explainability, which we will then use in later sections.
Then, \cref{sec:myth01,sec:myth02,sec:myth03} will disprove the myths
of XAI listed above.
\cref{sec:disc} briefly discusses the significance of disproving those
myths of XAI.
The paper concludes in~\cref{sec:conc}.

\section{Preliminaries} \label{sec:prelim}

\jnoteF{Classification Problems}

\paragraph{Classification in ML.}~
%
A classification problem is defined on a set of features
$\fml{F}=\{1,\ldots,m\}$, each with domain $\mbb{D}_i$, and a set of
classes $\fml{K}=\{c_1,c_2,\ldots,c_K\}$.
Domains can be categorical or ordinal, in this case with values that
can be integer- or real-valued.
Feature space $\mbb{F}$ is defined as the Cartesian product of the
domains of the features, in order:
$\mbb{F}=\mbb{D}_1\times\cdots\times\mbb{D}_m$.
We will use $\mbf{x}=(x_1,\ldots,x_m)$ to denote an arbitrary point in
feature space, i.e.\ each $x_i$ is a variable taking values from
$\mbb{D}_i$.
In contrast, we will use $\mbf{v}=(v_1,\ldots,v_m)$ to denote a
specific point in feature space, i.e.\ each $v_i$ is a constant taken
from $\mbb{D}_i$.
A classification function is a non-constant map from feature space
into the set of classes, $\kappa:\mbb{F}\to\fml{K}$.
(Clearly, a classifier would be useless if the classification function
were constant.)
A boolean classifier is one where the feature domains are boolean,
i.e.\ $\mbb{D}_i=\mbb{B}=\{0,1\}$, for $i\in\fml{F}$, and the set of
classes is also boolean, i.e.\ $\fml{K}=\mbb{B}$.
An instance is a pair $(\mbf{v},c)$ representing a point $\mbf{v}$ in
feature space, and the prediction by the classifier,
i.e.\ $\kappa(\mbf{v})=c$.
Finally, a classifier $\fml{M}$ is a tuple
$(\fml{F},\mbb{F},\fml{K},\kappa)$.
In addition, an explanation problem $\fml{E}$ is a tuple
$(\fml{M},(\mbf{v},c))$, where
$\fml{M}=(\fml{F},\mbb{F},\fml{K},\kappa)$ is a classifier.
Thus, in this paper, and given an explanation problem, our objective
is to define explanations and investigate approaches for computing
such explanations.

\paragraph{Sample of ML models.}~
%
In this paper we study so-called interpretable models, concretely
decision trees and decision sets, but also classifiers represented by
truth tables.
All these representations of classification functions are well-known,
and the concrete notation used is clarified with the running examples
presented in the rest of this section.

\paragraph{Running examples.}~
%
Throughout the paper, we will use a few simple running examples.
For the first parts of the paper, we will consider both a decision
tree (DT) and a decision list (DL). These rank among the simplest ML
models, and that simplicity often motivates claims of
interpretability.

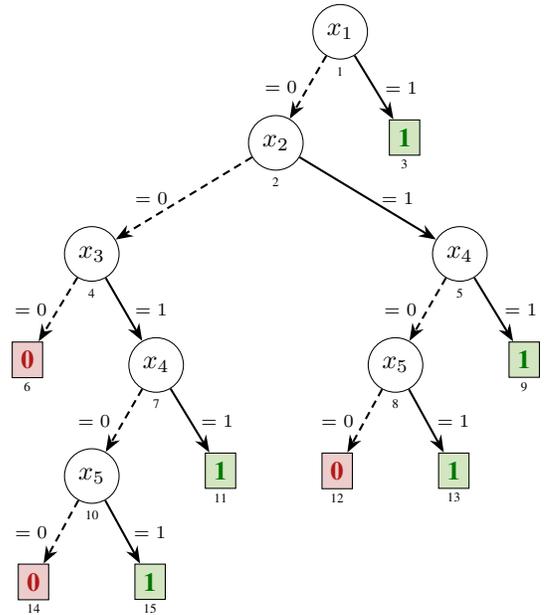
\begin{figure}[t]
  \begin{center}
    \scalebox{0.985}{
%
\forestset{
  BDT/.style={
    for tree={
      l=1.5cm,s sep=1.15cm,
      if n children=0{}{circle}, 
      draw=black,
      text=black,
      edge={
        my edge
      },
      if n=1{
        edge+={0 my edge},
      }{},
      edge=thick,
    }
  },
}
%
%
\begin{forest}
  BDT
  [{$x_1$}, label={[yshift=-6.875ex]{{\tiny1}}} 
    [{$x_2$}, label={[yshift=-6.875ex]{{\tiny2}}}, 
      edge label={node[midway,left,xshift=-0.5pt] {{\scriptsize$=0$}}}
      [{$x_3$}, label={[yshift=-6.875ex]{{\tiny4}}}, 
        edge label={node[midway,left,xshift=-2.5pt] {{\scriptsize$=0$}}}
        [\rhlight{\textbf{0}}, label={[yshift=-5.0ex]{{\tiny6}}},
          edge label={node[midway,left,xshift=-0.5pt] {{\scriptsize$=0$}}},
          rectangle, fill={tred3!20} ]
        [{$x_4$}, label={[yshift=-6.875ex]{{\tiny7}}}, 
          edge label={node[midway,right,xshift=0.5pt] {{\scriptsize$=1$}}}
          [{$x_5$}, label={[yshift=-6.875ex]{{\tiny10}}}, 
            edge label={node[midway,left,xshift=-1.5pt] {{\scriptsize$=0$}}}
            [\rhlight{\textbf{0}}, label={[yshift=-5.0ex]{{\tiny14}}},
              edge label={node[midway,left,xshift=-1.5pt] {{\scriptsize$=0$}}},
              rectangle, fill={tred3!20} ]
            [\dghlight{\textbf{1}}, label={[yshift=-5.0ex]{{\tiny15}}},
              edge label={node[midway,right,xshift=0.5pt] {{\scriptsize$=1$}}},
              rectangle, fill={tgreen3!25} ]
          ]
          [\dghlight{\textbf{1}}, label={[yshift=-5.0ex]{{\tiny11}}},
            edge label={node[midway,right,xshift=0.5pt] {{\scriptsize$=1$}}},
            rectangle, fill={tgreen3!25} ]
        ]
      ]
      [{$x_4$}, label={[yshift=-6.875ex]{{\tiny5}}}, 
        edge label={node[midway,right,xshift=1.5pt] {{\scriptsize$=1$}}}
        [{$x_5$}, label={[yshift=-6.875ex]{{\tiny8}}}, 
          edge label={node[midway,left,xshift=-0.5pt] {{\scriptsize$=0$}}}
          [\rhlight{\textbf{0}}, label={[yshift=-5.0ex]{{\tiny12}}},
            edge label={node[midway,left,xshift=-0.5pt] {{\scriptsize$=0$}}},
            rectangle, fill={tred3!20} ]
          [\dghlight{\textbf{1}}, label={[yshift=-5.0ex]{{\tiny13}}},
            edge label={node[midway,right,xshift=0.5pt] {{\scriptsize$=1$}}},
            rectangle, fill={tgreen3!25} ]
        ]
        [\dghlight{\textbf{1}}, label={[yshift=-5.0ex]{{\tiny9}}},
          edge label={node[midway,right,xshift=0.5pt] {{\scriptsize$=1$}}},
          rectangle, fill={tgreen3!25} ]
      ]
    ]
    [\dghlight{\textbf{1}}, label={[yshift=-5.0ex]{{\tiny3}}},
      edge label={node[midway,right,xshift=0.5pt] {{\scriptsize$=1$}}},
      rectangle, fill={tgreen3!25} ]
  ]
\end{forest}}
  \end{center}
  \caption{Example decision tree (adapted from~\cite{rudin-nips19})}
  \label{fig:dt}
\end{figure}

We consider the DT shown in~\cref{fig:dt}, adapted
from\cite{rudin-nips19}, and studied for explainability
in~\cite{iims-jair22,msi-fai23}. The nodes are numbered 1 to 15. The
internal nodes test the value of a (single) feature, and the terminal
nodes show the predicted class. Each edge defines a literal on the
feature associated with the edge's starting node.
Paths are sequences of nodes, starting at the root, and terminating at
a terminal node. Path $\langle1,2,4,6\rangle$ is an example of a path.
The literals associated with this path are: $x_1=0$, $x_2=0$ and
$x_3=0$.
Given an instance $(\mbf{v},c)=((0,0,0,0,0),0)$, the path consistent
with $\mbf{v}=(v_1,v_2,v_3,v_4,v_5)$ is
$\langle1,2,4,6\rangle$, where each $v_i$ is assigned to feature $i$.
Similarly, for the instance $((0,1,1,0,0),1)$ is consistent with path
$\langle1,2,5,8,12\rangle$.
As described elsewhere~\cite{rudin-nips19}, the DT was generated with
an algorithm for learning optimal DTs. Hence, the DT does not exhibit
any redundancy in terms a cost function defined on an aggregated set
of criteria, namely the loss (or misclassification error) and tree
size.

Moreover, we also consider the DL shown in~\cref{fig:dl}.
\begin{figure}[t]
  %
  %
  \[
  \begin{array}{lclclc}
    \tn{R}_1: & \quad &
    \tn{IF} & ({x_1}\land{x_3}) & \tn{THEN} & \kappa(\mbf{x})=1 \\
    \tn{R}_2: & \quad &
    \tn{ELSE IF} & (x_2\land{x_4}\land{x_6}) & \tn{THEN} & \kappa(\mbf{x})=0 \\
    \tn{R}_3: & \quad &
    \tn{ELSE IF} & (\neg{x_1}\land{x_3}) & \tn{THEN} & \kappa(\mbf{x})=1 \\
    \tn{R}_4: & \quad &
    \tn{ELSE IF} & (x_4\land{x_6}) & \tn{THEN} & \kappa(\mbf{x})=0 \\
    \tn{R}_5: & \quad &
    \tn{ELSE IF} & (\neg{x_1}\land\neg{x_3}) & \tn{THEN} &
    \kappa(\mbf{x})=1 \\
    \tn{R}_6: & \quad &
    \tn{ELSE IF} & (x_6) & \tn{THEN} & \kappa(\mbf{x})=0 \\
    \tn{R}_\tn{DEF}: & \quad &
    \tn{ELSE} & & & \kappa(\mbf{x})=1\\
  \end{array}
  \]
  \caption{Example of decision list (adapted from~\cite{msi-fai23})}
  \label{fig:dl}
\end{figure}
If the condition of some rule $\tn{R}_i$ is true, and the conditions
of previous rules are false, then the prediction is the one of rule
$\tn{R}_i$, and we say that that rule fires.
As an example, for the point in feature space
$(x_1,x_2,x_3,x_4,x_5,x_6)=(1,1,1,1,1,1)$, rule $\tn{R}_1$ fires since
$x_1=x_3=1$.


We will also consider two boolean classifiers represented by their
truth tables, as shown in~\cref{fig:bfunc}.
The instances that will be analyzed in more detail are highlighted,
and correspond to $((0,0,1,1),0)$ for~\cref{ex:k4}, and
$((1,1,1,1),0)$ for~\cref{ex:k5}.

\begin{figure}
  \begin{subfigure}{0.475\linewidth}
    \centering
    \renewcommand{\arraystretch}{0.95}
    \renewcommand{\tabcolsep}{0.5em}
    \scalebox{0.95}{
      \begin{tabular}[t]{ccccc} \toprule
        $x_1$ & $x_2$ & $x_3$ & $x_4$ & $\kappa_{I4}(\mbf{x})$ \\ \toprule
        0 & 0 & 0 & 0 & 0\\
        0 & 0 & 0 & 1 & 0\\
        0 & 0 & 1 & 0 & 0\\
        \tikzmarknode{e}{0} & 0 & 1 & 1 & \tikzmarknode{f}{0}\\
        0 & 1 & 0 & 0 & 0\\
        0 & 1 & 0 & 1 & 0\\
        0 & 1 & 1 & 0 & 0\\
        0 & 1 & 1 & 1 & 1\\
        1 & 0 & 0 & 0 & 0\\
        1 & 0 & 0 & 1 & 0\\
        1 & 0 & 1 & 0 & 1\\
        1 & 0 & 1 & 1 & 0\\
        1 & 1 & 0 & 0 & 1\\
        1 & 1 & 0 & 1 & 1\\
        1 & 1 & 1 & 0 & 1\\
        1 & 1 & 1 & 1 & 1\\
        \bottomrule
      \end{tabular}
    }
    \caption{Function $\kappa_{I4}$} \label{ex:k4}
    \begin{tikzpicture}[overlay,remember picture]
      \node[draw=midblue, thin, xshift=-1.5pt, yshift=2.75pt, inner sep=1.5pt, fit=(e) (f)] {};
    \end{tikzpicture}
  \end{subfigure}
  \begin{subfigure}{0.475\linewidth}
    \centering
    \renewcommand{\arraystretch}{0.95}
    \renewcommand{\tabcolsep}{0.5em}
    \scalebox{0.95}{
      \begin{tabular}[t]{ccccc} \toprule
        $x_1$ & $x_2$ & $x_3$ & $x_4$ & $\kappa_{I5}(\mbf{x})$ \\ \toprule
        0 & 0 & 0 & 0 & 0\\
        0 & 0 & 0 & 1 & 0\\
        0 & 0 & 1 & 0 & 0\\
        0 & 0 & 1 & 1 & 0\\
        0 & 1 & 0 & 0 & 0\\
        0 & 1 & 0 & 1 & 0\\
        0 & 1 & 1 & 0 & 0\\
        0 & 1 & 1 & 1 & 1\\
        1 & 0 & 0 & 0 & 0\\
        1 & 0 & 0 & 1 & 0\\
        1 & 0 & 1 & 0 & 0\\
        1 & 0 & 1 & 1 & 1\\
        1 & 1 & 0 & 0 & 0\\
        1 & 1 & 0 & 1 & 1\\
        1 & 1 & 1 & 0 & 0\\
        \tikzmarknode{g}{1} & 1 & 1 & 1 & \tikzmarknode{h}{0}\\
        \bottomrule
      \end{tabular}
    }
    \caption{Function $\kappa_{I5}$} \label{ex:k5}
    \begin{tikzpicture}[overlay,remember picture]
      \node[draw=midblue, thin, xshift=-1.25pt, yshift=9.5pt, inner sep=1.5pt, fit=(g) (h)] {};
    \end{tikzpicture}
  \end{subfigure}
  \caption{Examples of boolean classifiers} \label{fig:bfunc}
\end{figure}

\section{A Glimpse of Explainable AI} \label{sec:xai}

\subsection{Intrinsic Interpretability}
\label{ssec:ii}

\jnoteF{Use so-called interpretable models, especially in high-risk
  domains.}

Intrinsic interpretability~\cite{molnar-bk20} builds on a long-held
  belief that some ML models are naturally interpretable. (This belief
  has been stated for example for decision trees~\cite{breiman-ss01},
  but also for decision sets~\cite{leskovec-kdd16}, among others.)
Indeed, it is the case that, for some ML models, an explanation can be
produced by some sort of manual inspection of the model. This is the
case with decision trees, decision lists, decision sets, but other
graph-based ML models~\cite{hiims-kr21}.

Recent years have witnessed proposals for the use of interpretable
models in high-risk
applications~\cite{rudin-naturemi19,rudin-naturermp22}.
Unfortunately, it is unclear that
a rigorous definition of interpretability can be
devised~\cite{lipton-cacm18}. Given the motivation for work on
interpretable models, we equate interpretability with succinctness and
irreducibility of explanations.

\subsection{Feature-Based Explanations}
\label{ssec:fbxp}

\jnoteF{Feature attribution vs.\ feature selection}

Most practical work on explainability relates explanations with
the model's features. Two main lines of work exist: \emph{feature
attribution} and \emph{feature selection}.

Feature attribution 
(e.g.~\cite{muller-plosone15,guestrin-kdd16a,lundberg-nips17}) assigns
some value of importance to each of the model's features. Relative
feature importance is then obtained by comparing the value of
importance of the different features.
Among others, Shapley values have found widespread acceptance as a
mechanism for feature attribution in explainability.

Feature selection
(e.g.~\cite{guestrin-kdd16a,darwiche-ijcai18,inms-aaai19}) identifies
sets of features that are sufficient (probabilistically or
deterministically) for the prediction. The most visible feature
selection methods are heuristic, but there has been recent work on
formal approaches for feature selection in explainability; this is
discussed in~\cref{ssec:fxai}.

\subsection{Non-Formal Explainability}
\label{ssec:nfxai}

The vast majority of work on explainability has been based on what we
refer to as \emph{non-formal explainability}, i.e.\ approaches that
offer no guarantees of rigor in computed explanations.
There are fairly up to date overviews of non-formal explainability
approaches~\cite{muller-dsp18,pedreschi-acmcs19,xai-bk19,muller-xai19-ch01,molnar-bk20,muller-ieee-proc21}.

One well-known line of work are model-agnostic
methods~\cite{guestrin-kdd16a,lundberg-nips17,guestrin-aaai18}.
Model-agnostic methods approach explainability by ignoring the actual
implementation of the ML model, and either sample the behavior from
training data, or sample the black-box behavior of the model using an
inferred distribution from training data.
As an example, SHAP~\cite{lundberg-nips17} approaches feature
attribution by proposing to approximate the computation of Shapley
values for explainability.

\jnoteF{Model-agnostic vs. feature attribution methods, e.g. in NNs}

\jnoteF{Cite example surveys.}

\subsection{Shapley Values for Explainability}
\label{ssec:svs}

\jnoteF{SVs as proposed in earlier work, i.e.\ paper of Barcelo}

Shapley values were first introduced by
L.~Shapley~\cite{shapley-ctg53} in the context of game theory.
Shapley values have been extensively used for explaining the
predictions of ML models,
e.g.~\cite{kononenko-jmlr10,kononenko-kis14,zick-sp16,lundberg-nips17,jordan-iclr19,taly-cdmake20,lakkaraju-nips21,watson-facct22},
among a vast number of recent examples.
The complexity of computing Shapley values (as proposed in
SHAP~\cite{lundberg-nips17}) has been studied in recent
years~\cite{barcelo-aaai21,vandenbroeck-aaai21,barcelo-corr21,vandenbroeck-jair22,cms-aij23}.
This section provides a brief overview of Shapley values. 
Throughout the section, we adapt the notation used in recent
work~\cite{barcelo-aaai21,barcelo-corr21}, which builds on the work
of~\cite{lundberg-nips17}.

Let $\Upsilon:2^{\fml{F}}\to2^{\mbb{F}}$ be defined by,
\begin{equation} \label{eq:upsilon}
  \Upsilon(\fml{S};\mbf{v})=\{\mbf{x}\in\mbb{F}\,|\,\land_{i\in\fml{S}}x_i=v_i\}
\end{equation}
i.e.\ for a given set $\fml{S}$ of features, and parameterized by
the point $\mbf{v}$ in feature space, $\Upsilon(\fml{S};\mbf{v})$
denotes all the points in feature space that have in common with
$\mbf{v}$ the values of the features specified by $\fml{S}$. 

Also, let $\phi:2^{\fml{F}}\to\mbb{R}$ be defined by,
\begin{equation} \label{eq:phi}
  \phi(\fml{S};\fml{M},\mbf{v})=\frac{1}{2^{|\fml{F}\setminus\fml{S}|}}\sum_{\mbf{x}\in\Upsilon(\fml{S};\mbf{v})}\kappa(\mbf{x})
\end{equation}
For the purposes of this paper, and in contrast
with~\cite{barcelo-aaai21,barcelo-corr21}, one can solely consider a
uniform input distribution, and so the dependency with the input
distribution is not accounted for.
As a result, 
given a set $\fml{S}$ of features, $\phi(\fml{S};\fml{M},\mbf{v})$
represents the average value of the classifier over the points of
feature space represented by $\Upsilon(\fml{S};\mbf{v})$.

Finally, let $\sv:\fml{F}\to\mbb{R}$ be defined by,
\begin{align} \label{eq:sv}
  \sv(i;\fml{M},\mbf{v})=&\sum_{\fml{S}\subseteq(\fml{F}\setminus\{i\})}\frac{|\fml{S}|!(|\fml{F}|-|\fml{S}|-1)!}{|\fml{F}|!}\times\nonumber\\[5pt]
  &\left(\phi(\fml{S}\cup\{i\};\fml{M},\mbf{v})-\phi(\fml{S};\fml{M},\mbf{v})\right)
\end{align}
Given an instance $(\mbf{v},c)$, the Shapley values assigned to each
feature measure the \emph{contribution} of the features with respect
to the prediction. 
A positive/negative value indicates that the feature can contribute to
changing the prediction, whereas a value of 0 indicates no
contribution.
Moreover, and as our results demonstrate, SHAP never really replicates
exact Shapley values. As a result, we focus on the relative order of
features imposed by the computed Shapley values. The motivation is
that, even if the computed values are not correct (as in the case of
SHAP), then what matters for a human decision maker is the order of
features.

\jnoteF{To
  cite:~\cite{shapley-ctg53,barcelo-aaai21,vandenbroeck-aaai21,vandenbroeck-jair22}.}

\begin{example} \label{ex:calcsv}
  We consider the example boolean functions of~\cref{fig:bfunc}.
  If the functions are represented by a truth table, then the Shapley
  values can be computed in polynomial time on the size of the truth
  table, since the number of subsets considered in~\eqref{eq:sv} is
  also polynomial on the size of the truth table~\cite{hms-corr23}. 
  (Observe that for each subset used in~\eqref{eq:sv}, we can
  use the truth table for computing the average values
  in~\eqref{eq:phi}.)
  For example, for $\kappa_{I4}$ (see~\cref{ex:k4})
  and for the point in feature space
  $(0,0,1,1)$, one can compute the following Shapley values:
  $\sv(1)=-0.125$,  $\sv(2)=-0.333$, $\sv(3)=0.083$, and $\sv(4)=0.0$.
\end{example}

\subsection{Formal Explainability}
\label{ssec:fxai}

This section offers a brief perspective of the emerging field of
formal explainable AI. Recent detailed accounts are
available~\cite{msi-aaai22,ms-corr22}.

\paragraph{Abductive explanations (AXp's).}~
%
An AXp~\cite{inms-aaai19}\footnote{%
AXp's are also referred to as prime implicant (PI)
explanations~\cite{darwiche-ijcai18} and as sufficient reasons for a
prediction.}
represents a subset-minimal set of literals (relating a feature value 
$x_i$ and a constant $v_i\in\mbb{D}_i$) that are logically sufficient
for the prediction. AXp's can be viewed as answering a \textbf{Why?}
question, i.e.\ why the prediction.
(AXp's can be described in terms of logic-based abduction, which
explains the name used~\cite{msi-aaai22}.)
AXp's offer guarantees of rigor that are not offered by other
alternative explanation approaches.
%
More recently, AXp's have been studied
in terms of their computational
complexity~\cite{barcelo-nips20,marquis-kr21,marquis-dke22,darwiche-jlli23}.
There is a growing body of recent work on formal
explanations~\cite{darwiche-jair21,barcelo-nips21,kutyniok-jair21,kwiatkowska-ijcai21,mazure-cikm21,tan-nips21,rubin-aaai22,msi-aaai22,amgoud-ijcai22,leite-kr22,barcelo-corr22,lorini-jlc23}.
Formally, given $\mbf{v}=(v_1,\ldots,v_m)\in\mbb{F}$, with
$\kappa(\mbf{v})=c$, an AXp is any subset-minimal set
$\fml{X}\subseteq\fml{F}$ such that\footnote{%
In the remainder of this section, the parameterization of the presented
predicates on the classifier $\fml{M}$ and on $(\mbf{v},c)$ is omitted
for the sake of brevity.},
\begin{align} \label{eq:axp}
  \waxp(&\fml{X}) {~:=~} \nonumber \\
  & \forall(\mbf{x}\in\mbb{F}).
  \left[
    \bigland_{i\in{\fml{X}}}(x_i=v_i) 
    \right]
  \limply(\kappa(\mbf{x})=c)
\end{align}
If a set $\fml{X}\subseteq\fml{F}$ is not 
minimal but \eqref{eq:axp} holds, then $\fml{X}$ is referred to as a
\emph{weak} AXp.
%
%
%
Clearly, the predicate $\waxp$ maps $2^{\fml{F}}$ into $\{\bot,\top\}$
(or $\{\False,\True\}$).
Given $\mbf{v}\in\mbb{F}$, an AXp $\fml{X}$ represents an irreducible
(or minimal) subset of the features which, if assigned the values
dictated by $\mbf{v}$, are sufficient for the prediction $c$,
i.e.\ value changes to the features not in $\fml{X}$ will not change
the prediction.
We can use the definition of the predicate $\waxp$ to formalize the
definition of the predicate $\axp$, also defined on subsets $\fml{X}$
of $\fml{F}$:
\begin{align} \label{eq:axp2}
  \axp(&\fml{X}) {~:=~} \nonumber \\
  & \waxp(\fml{X}) \land
  \forall(\fml{X}'\subsetneq\fml{X}).
  \neg\waxp(\fml{X}')
\end{align}
The definition of $\waxp(\fml{X})$ ensures that the predicate is 
\emph{monotone}.
Indeed, if $\fml{X}\subseteq\fml{X}'\subseteq\fml{F}$, and if
$\fml{X}$ is a weak AXp, then $\fml{X}'$ is also a weak AXp, as the
fixing of more features will not change the prediction.
%
%
%
Since the $\waxp$ predicate is monotonic, the definition of the
predicate $\axp$ can be simplified as follows, with
$\fml{X}\subseteq\fml{F}$: 
\begin{align} \label{eq:axp3}
  \axp(&\fml{X}) {~:=~} \nonumber \\
  & \waxp(\fml{X}) \land
  \forall(j\in\fml{X}).\neg\waxp(\fml{X}\setminus\{j\})
\end{align}
This simpler but equivalent definition of AXp has important practical
significance, in that only a linear number of subsets needs to be
checked for, as opposed to exponentially many subsets
in~\eqref{eq:axp2}. As a result, the algorithms that compute one AXp
are based on~\eqref{eq:axp3}~\cite{msi-aaai22}.

\begin{example} \label{ex:axp}
  For the boolean function $\kappa_{I4}$ in~\cref{ex:k4}, and the
  instance $((0,0,1,1),0)$, we can observe that the value of the
  function is 0 for the first four rows of the truth table. Thus, if
  both $x_1$ and $x_2$ equal 1, then the prediction is guaranteed to
  be 0. Hence, $\{1,2\}$ is a weak AXp. By inspection of the remaining
  rows, where either $x_1$ or $x_2$ changes, we can conclude that
  $\{1,2\}$ is indeed an AXp.
\end{example}

\paragraph{Contrastive explanations (CXp's).}~
%
A CXp is a subset-minimal set of features which suffices to change the
prediction.

Similarly to the case of AXp's, one can define (weak) contrastive
explanations (CXp's)~\cite{miller-aij19,inams-aiia20}.
$\fml{Y}\subseteq\fml{F}$ is a weak CXp for the instance $(\mbf{v},c)$
if,
\begin{align} \label{eq:cxp1}
  \wcxp(&\fml{Y}) ~{:=}~ \nonumber\\
  & \exists(\mbf{x}\in\mbb{F}).%
    \left[\bigwedge_{i\not\in\fml{Y}}(x_i=v_i)\right]\land(\kappa(\mbf{x})\not=c)
\end{align}
%
%
Thus, given an instance $(\mbf{v},c)$, a (weak) CXp is a subset of
features which, if allowed to take any value from their domain, then
there is an assignment to the features that changes the prediction to
a class other than $c$, this while the features not in the explanation
are kept to their values. 
%

Furthermore, a set $\fml{Y}\subseteq\fml{F}$ is a CXp if, besides
being a weak CXp, it is also subset-minimal, i.e.
\begin{align} \label{eq:cxp2}
  \cxp(&\fml{Y}) ~{:=}~ \nonumber\\
  & \wcxp(\fml{Y})\land\forall(\fml{Y}'\subsetneq\fml{Y}).\neg\wcxp(\fml{Y}')
\end{align}  
A CXp can be viewed as a possible irreducible answer to a \tbf{Why
  Not?} question, i.e.\ why isn't the classifier's prediction a
class other than $c$?

As before, monotonicity of predicate $\wcxp$ allows
simplifying~\cref{eq:cxp2} to get,
\begin{align} \label{eq:cxp3}
  \cxp(&\fml{Y}) {~:=~} \nonumber \\
  & \wcxp(\fml{Y}) \land
  \forall(j\in\fml{Y}).\neg\wcxp(\fml{Y}\setminus\{j\})
\end{align}

\begin{example} \label{ex:cxp}
  For the example function~$\kappa_{I4}$, and instance
  $((0,0,1,1),0)$, if we fix features 1, 3 and 4, respectively to 0, 1
  1, then by allowing feature 2 to change value, we see that the
  prediction changes, e.g.\ by considering the point $(0,1,1,1)$ with
  prediction 1. Thus, $\{2\}$ is a CXp.
  In a similar way, by fixing the features 2 and 3, respectively to 0
  and 1, then by allowing features 1 and 4 to change value, we
  conclude that the prediction changes. Hence, $\{1,4\}$ is also a
  CXp.
\end{example}


%
%
%

\paragraph{Duality between explanations.}~
%
%
%
Given an explanation problem $\fml{E}$, the sets of AXp's and CXp's
are defined as follows:
\begin{equation} \label{eq:allxp}
  \begin{array}{l}
    \mbb{A}(\fml{E})=\{\fml{X}\subseteq\fml{F}\,|\,\axp(\fml{X})\}\\[3pt]
    \mbb{C}(\fml{E})=\{\fml{Y}\subseteq\fml{F}\,|\,\cxp(\fml{Y})\}
  \end{array}
\end{equation}

By taking into account the definition of AXp, CXp, and their sets
(see~\eqref{eq:allxp}), and also by building on Reiter's seminal 
work~\cite{reiter-aij87}, recent work~\cite{inams-aiia20} proved the 
following duality between minimal hitting sets (MHSes)\footnote{%
Given a set $\fml{S}$ of sets, an MHS is a subset-minimal set of
elements such that the intersection with each of the sets of $\fml{S}$
is non-empty.}:
\begin{prop}[Minimal hitting-set (MHS) duality between AXp's and CXp's]
  \label{prop:xpdual}
  Given an explanation problem $\fml{E}$, we have that,
  \begin{enumerate}
  \item $\fml{X}\subseteq\fml{F}$ is an AXp (and so
    $\fml{X}\in\mbb{A}(\fml{E})$)
    iff $\fml{X}$ is an MHS of the CXp's in $\mbb{C}(\fml{E})$.
  \item $\fml{Y}\subseteq\fml{F}$ is a CXp (and so
    $\fml{Y}\in\mbb{C}(\fml{E})$)
    iff $\fml{Y}$ is an MHS of the AXp's in $\mbb{A}(\fml{E})$.
  \end{enumerate}
\end{prop}
We refer to~\cref{prop:xpdual} as MHS duality between AXp's and CXp's.
The previous result has been used in more recent papers for enabling the
enumeration of
explanations~\cite{msgcin-icml21,ims-sat21,hiims-kr21}.

\begin{example}
    For the function in~\cref{ex:k5}, and instance $((1,1,1,1),0)$,
    there is only one AXp: $\mbb{A}(\fml{E})=\{\{1,2,3\}\}$,
    i.e.\ $\{1,2,3\}$ is the only subset-minimal set of features
    which, if fixed to the values dictated by $\mbf{v}$, is sufficient
    for predicting 0. Moreover, the set of CXp's is:
    $\mbb{C}(\fml{E})=\{\{1\},\{2\},\{3\}\}$. Clearly, each CXp is an
    MHS of the set AXp's and the AXp is an MHS of the set of CXp's.
\end{example}

Let us defined the sets
$F_{\mbb{A}}(\fml{E})=\cup_{\fml{X}\in\mbb{A}(\fml{E})}\fml{X}$ and
$F_{\mbb{C}}(\fml{E})=\cup_{\fml{Y}\in\mbb{C}(\fml{E})}\fml{Y}$.
$F_{\mbb{A}}(\fml{E})$ aggregates the features occurring in any
abductive explanation, whereas $F_{\mbb{C}}(\fml{E})$ aggregates the
features occurring in any contrastive explanation.
%
%
%
Given the above, the following results are a consequence of
\cref{prop:xpdual}:
\begin{prop} \label{prop:dual2}
  ${F}_{\mbb{A}}(\fml{E})={F}_{\mbb{C}}(\fml{E})$.
\end{prop}

\begin{cor} \label{prop:xpdual2}
  Given a classifier function $\kappa:\mbb{F}\to\fml{K}$, defined on a
  set of features $\fml{F}$, a feature $i\in\fml{F}$ is
  included in some AXp iff $i$ is included in some CXp.
\end{cor}

Additional duality results in (formal) explainability have been
established elsewhere~\cite{inms-nips19}.

\paragraph{Explanability queries -- feature relevancy~\cite{hiims-kr21,hims-aaai23,hcmpms-tacas23}.}~
%
Let us consider a classifier $\fml{M}$, with features $\fml{F}$,
domains $\mbb{D}_i$, $i\in\fml{F}$, classes $\fml{K}$, a
classification function $\kappa:\mbb{F}\to\fml{K}$,  and a concrete
instance $(\mbf{v},c)$, $\mbf{v}\in\mbb{F},c\in\fml{K}$.

\begin{defn}[Feature Necessity, Relevancy \& Irrelevancy]
  Let $\mbb{A}$ denote the set of all AXp's for a classifier given
  a concrete instance, i.e.
  $\mbb{A} = \{\fml{X}\subseteq\fml{F}\,|\,\axp(\fml{X})\}$,
  %
  and let $t\in\fml{F}$ be a target feature.
  Then,
  \begin{enumerate}[nosep,label=\roman*.] 
  \item $t$ is necessary if $t\in\cap_{\fml{X}\in\mbb{A}}\fml{X}$;
  \item $t$ is relevant if $t\in\cup_{\fml{X}\in\mbb{A}}\fml{X}$; and
  \item $t$ is irrelevant if
    $t\in\fml{F}\setminus\cup_{\fml{X}\in\mbb{A}}\fml{X}$.
  \end{enumerate}
\end{defn}
%
%
Because of feature occurs in some AXp iff it occurs in some CXp, then
the previous definitions could consider instead the set of contrastive
explanations.

\begin{example} \label{ex:frp}
  For the example function ~\cref{ex:k5}, let us consider the instance
  $((1,1,1,1),0)$.
  By inspection, we conclude that the only AXp is $\{1,2,3\}$.
  Hence, features $1,2,3$ are relevant, and feature $4$ is
  irrelevant.,
\end{example}

In the rest of the paper, we will use the predicate $\relevant(i)$ to
hold true when feature $i$ is relevant, and the predicate
$\irrelevant(i)$ to hold true when feature $i$ is irrelevant.

\begin{figure}[t]
  \centering \begin{tikzpicture}
[
  good/.style={
    draw=tgreen3,
    thick,
    fill=tgreen3!35,
    minimum width=0.5cm,
    minimum height=0.5cm,
    rounded corners=1mm,
    font=\tiny\sffamily\bfseries
  },
  bad/.style={
    draw=tred3,
    thick,
    fill=tred3!27,
    minimum width=0.5cm,
    minimum height=0.5cm,
    rounded corners=1mm,
    font=\tiny\sffamily\bfseries
  },
  soso/.style={
    draw=tblue3,
    thick,
    fill=tblue3!27,
    minimum width=0.5cm,
    minimum height=0.5cm,
    rounded corners=1mm,
    font=\tiny\sffamily\bfseries
  }
]

\def\a{1.9}
\def\b{1.9}
\path
(-3,-3.25) node[good]{NBCs~\cite{msgcin-nips20}}
(-1.475,-3.25) node[good]{DTs~\cite{iims-corr20,iims-corr22}}
(-3,-2.575) node[good]{XpGs~\cite{hiims-kr21}}
(-1.65,-2.575) node[good]{GDFs~\cite{hiims-kr21}}
(-2.625,-1.9) node[good]{Monotonic~\cite{msgcin-icml21,cms-cp21}}
(-1.425,-1.225) node[good]{d-DNNF~\cite{hiicams-aaai22}}

(-2.9,0.5) node[soso]{DLs~\cite{ims-sat21}}
(-2.3,1.15) node[soso]{RFs~\cite{ims-ijcai21}}
(-1.5,1.8) node[soso]{BTs~\cite{ignatiev-ijcai20,iisms-aaai22}}

(2,2.075) node[bad]{NNs~\cite{inms-aaai19}}
(2.8,3.05) node[bad]{BNs~\cite{darwiche-ijcai18}}
;
\draw[dotted,thick] (2*\a,0)--(-2*\a,0) (0,2*\b)--(0,-2*\b);
\draw[solid,thick] (-2*\a,-2*\b)--+(0:4*\a);
\draw[solid,thick] (-2*\a,2*\b)--+(0:4*\a);
\draw[solid,thick] (-2*\a,-2*\b)--+(90:4*\b);
\draw[solid,thick] (2*\a,-2*\b)--+(90:4*\b);
\path
(0,-4.5) node{\small Practical scalability (effectiveness)}
(-3,-4.075) node{\scriptsize Effective}
(3,-4.0075) node{\scriptsize Ineffective}
(-4.5,0) node[rotate=90]{\small Computational complexity}
(-4.075,1.8) node[rotate=90]{\scriptsize Computationally hard}
(-4.075,-1.8) node[rotate=90]{\scriptsize Poly-time}
(0,4.075) node{\small Computing one XP}
;
\end{tikzpicture}

  \caption{Progress in formal XAI~\cite{msi-aaai22,ms-corr22}}
  \label{fig:progress}
\end{figure}
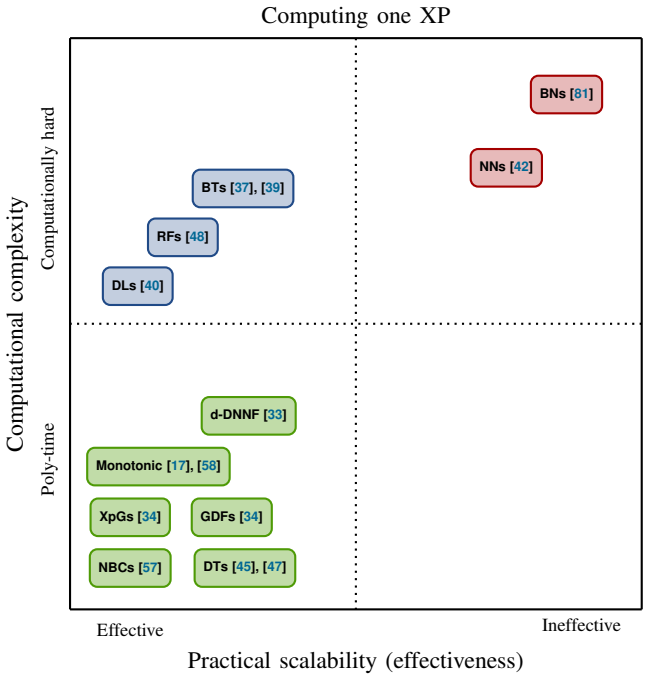

\paragraph{Progress in formal XAI.}~
%
\cref{fig:progress} summarizes the progress observed in recent years
for computing one explanation~\cite{msi-aaai22,ms-corr22}.
For several families of classifiers there exist practical efficient
solutions, either in theory or in practice. These include Naive Bayes
Classifiers~\cite{msgcin-nips20}, Decision
Trees~\cite{iims-corr20,iims-jair22}, restricted boolean circuits
(e.g.\ d-DNNFs, GDFs, BDDs)~\cite{hiims-kr21,hiicams-aaai22} and
monotonic classifiers~\cite{msgcin-icml21,cms-cp21}, but also decision
lists~\cite{ims-sat21}, random forests and boosted
trees~\cite{ignatiev-ijcai20,ims-ijcai21,iisms-aaai22}.
For some families of classifiers, e.g.\ Neural Networks and Bayesian
Network Classifiers, scalability is still an
issue~\cite{darwiche-ijcai18,inms-aaai19}.

Recent overviews of formal XAI 
cover additional detail~\cite{msi-aaai22,ms-corr22}, including recent
work on different explainability queries.
Furthermore, and with the purpose of disproving one of the myths in
XAI, recent work~\cite{hms-corr23} demonstrated that for classifiers
represented with truth tables, there exist polynomial-time algorithm
for computing one or more explanations (both abductive and
contrastive), but also for answering feature relevancy queries.

\jnoteF{Definition of AXp, CXp, duality between both.}

\jnoteF{Progress in FXAI in recent years -- show figure of progress.}

\jnoteF{Explainability queries: feature necessity \& relevancy}

\jnoteF{Cite recent surveys}

\section{Myth \#01: Intrinsic Interpretability}
\label{sec:myth01}

One approach to XAI is \emph{intrinsic interpretability}~\cite{molnar-bk20},
based on a long-held belief that some ML models are
\emph{interpretable}. Concrete examples include decision trees, lists
and sets. In the case of decision trees, their interpretability has
been asserted for more than two decades~\cite{breiman-ss01}.
The assumed interpretability of some families of ML models has
justified recent calls for their use 
in high-risk domains~\cite{rudin-naturemi19,rudin-naturermp22}.

\paragraph{Decision trees.}~
%
Given a DT and an instance $(\mbf{v},c)$, an intuitive explanation for
the prediction are the literals tested in the path consistent with
$\mbf{v}$.

\begin{example}
  For the DT of~\cref{fig:dt}, and the instance
  $(\mbf{v},c)=((0,0,0,0,0),0)$, the intuitive explanation for the
  prediction are the features tested in the path $\langle1,2,4,6\}$,
  and so we could propose as an explanation for why the prediction the
  set of literals $\{1,2,3\}$.
\end{example}

However, as proved in earlier
work~\cite{iims-corr20,hiims-kr21,iims-jair22}, such approach for
computing explanations in DTs can provide arbitrarily redundant
results.

\begin{example}
  Let us consider again the DT of~\cref{fig:dt}, but now let the
  target instance be $(\mbf{v},c)=(0,0,1,0,1),1)$.
  Following the approach outlined above, a possible explanation would
  be $\{1,2,3,4,5\}$, representing the path consistent with the
  instance.
  However, it is possible to prove that there exists a significantly
  simpler abductive explanation. If we pick the features $\{3,5\}$ as
  a possible AXp, it is simple to conclude that, for any assignment to
  the features 1, 2, and 4, the prediction never changes, as shown
  in~\cref{tab:dt1}. Hence, $\{3,5\}$ is a WAXp, and it is simple to
  conclude that it is indeed an AXp (i.e.\ any subset of $\{3,5\}$ is
  no longer a WAXp).
\end{example}

\begin{table}[t]
  \begin{center}
    \begin{tabular}{ccccc} \toprule
      $x_3=x_5=1$ & $x_1$ & $x_2$ & $x_4$ & $\kappa_{\tn{DT}}(\mbf{x})$ \\
      \toprule
      1 & 0 & 0 & 0 & 1 \\
      1 & 0 & 0 & 1 & 1 \\
      1 & 0 & 1 & 0 & 1 \\
      1 & 0 & 1 & 1 & 1 \\
      1 & 1 & 0 & 0 & 1 \\
      1 & 1 & 0 & 1 & 1 \\
      1 & 1 & 1 & 0 & 1 \\
      1 & 1 & 1 & 1 & 1 \\
      \bottomrule
    \end{tabular}
  \end{center}
  \caption{DT classification given $x_3=x_5=1$} \label{tab:dt1}
\end{table}

\cref{tab:dtred} summarizes the explanation redundancy of the paths in
the DT of~\cref{fig:dt}. Although some paths exhibit no redundancy,
i.e.\ the AXp corresponds to the path features, there are other paths
with up to 60\% redundant features.

\begin{table}[t]
  \begin{center}
    \renewcommand{\arraystretch}{1.15}
    \begin{tabular}{lccc} \toprule
      Path & Features & AXp & \%Red\\ \toprule
      $\langle1,2,4,6\rangle$ & $\{1,2,3\}$ & $\{1,2,3\}$ & 0\%\\
      $\langle1,2,4,7,10,14\rangle$ & $\{1,2,3,4,5\}$ & $\{1,4,5\}$ & 40\%\\
      $\langle1,2,4,7,10,15\rangle$ & $\{1,2,3,4,5\}$ & $\{3,5\}$ & 60\%\\
      $\langle1,2,4,7,11\rangle$ & $\{1,2,3,4\}$ & $\{3,4\}$ & 50\%\\
      $\langle1,2,5,8,12\rangle$ & $\{1,2,4,5\}$ & $\{1,4,5\}$ & 40\%\\
      $\langle1,2,5,8,13\rangle$ & $\{1,2,4,5\}$ & $\{2,5\}$ & 60\% \\
      $\langle1,2,5,9\rangle$ & $\{1,2,4\}$ & $\{2,4\}$ & 33\%\\
      $\langle1,3\rangle$ & $\{1\}$ & $\{1\}$ & 0\%\\
      \bottomrule
    \end{tabular}
  \end{center}
  \caption{Explanation redundancy figures for DT in~\cref{fig:dt}}
  \label{tab:dtred}
\end{table}

Besides comprehensive experimental evidence illustrating the
importance of explaining DTs~\cite{iims-jair22,msi-fai23}, recent work
proved a number of additional results.
First,~\cite{iims-corr20,iims-jair22} prove that
redundancy of path explanations can be arbitrarily large on the number
of features, i.e.\ there are DTs for which a path contains all the
features, and the explanation contains a single feature.
Second,~\cite{iims-jair22} proves that the family of classification
functions that do not exhibit path redundancy is very restricted.
The results above support the conclusions that DTs ought not serve as
explanations, but should be explained instead.

\paragraph{Decision lists.}~
%
For DLs it has also be argued that explanations are also
required~\cite{msi-fai23}, both to avoid redundancy of features in
explanations, but also to ensure the correctness of computed sets of
literals as explanations.

\begin{example}
  For the DL of~\cref{fig:dl}, let the instance be
  $((0,1,0,1,0,1),0)$. It is plain that the rule $\tn{R}_2$ fires, and
  the prediction is 0. What is a sufficient reason for this
  prediction? Clearly, $\{2,4,6\}$, as listed in the condition for the
  rule to fire, is not sufficient for the prediction.
  Indeed, we must ensure that rule $\tn{R}_1$ does not fire, and so
  the abductive explanation must either include feature 1 or feature
  3. Let us pick feature 3. It is the case that $\{2,3,4,6\}$ is a
  sufficient reason for the prediction given the values in $\mbf{v}$
  dictating the literals in features 2, 3, 4 and 6.
  However, one can prove that feature 2 is irrelevant. Indeed, if
  feature 3 is assigned value 0, and features 4 and 6 are assigned
  value 1, then because of rule $\tn{R}_4$, the prediction is
  guaranteed to be 0, independently of the value of feature 2. \\
  Clearly, if one of the last rules were to fire, finding an
  explanation would be even less intuitive than in the case above.
\end{example}

There is recent comprehensive practical evidence to the existence of
explanation redundancy in DLs~\cite{msi-fai23}.

It should be noted that explanation redundancy does not yield
incorrect explanations. However, redundancy of explanations is by
itself problematic. First, because it may cause human decision makers
to look at features that are unneeded for the prediction. Second,
because the number of concepts that human decision makers can relate
with is fairly small~\cite{miller-pr56}, and redundant information can
push explanations beyond that bound.

\jnoteF{To cite: \cite{breiman-ss01,rudin-naturemi19}...}

\jnoteF{Existing claims \& myths -- emphasize dramatic calls for the
  use of interpretable models}

\jnoteF{Discuss example(s): DT \& DL}

\jnoteF{Results that have been proved; \& current status}

\jnoteF{Plan:
  \begin{enumerate}[nosep]
  \item Some models have long been deemed interpretable: DTs, DLs,
    DSs. (But this assumes simple models.)
  \item A first question: do path explanations represent accurate
    reasons for a prediction?
  \item What we have proved: path explanations can be arbitrarily
    redundant on the number of features, i.e.\ we can have paths with
    all features, and a sufficient reason with size 1.
  \item How serious is this issue? We show that path explanation
    redundancy is ubiquitous, and so the use of path explanations to
    explain predictions is essentially guaranteed to yield redundant
    information.
  \end{enumerate}
}

\section{Myth \#02: Model-Agnostic Explainability}
\label{sec:myth02}

Model-agnostic explainability is arguably one of the most widely used
XAI approaches~\cite{guestrin-aaai18,lundberg-nips17,guestrin-kdd16a}.

One concrete line of work~\cite{guestrin-aaai18} finds a set of
features that ``explains'' the prediction. Concretely, an
\emph{anchor} is defined to be: ``\emph{An anchor explanation is a rule
that sufficiently 'anchors' the prediction locally -- such that
changes to the rest of the feature values of the instance do not
matter.}''~\cite{guestrin-aaai18}.

If the sets of features reported as anchors are indeed
\emph{sufficient} for the prediction, such that the other features do
not matter, then one should be able to validate those sets of features
as abductive explanations, since AXp's also \emph{anchor} the
prediction locally.

Given an instance, $(\mbf{v},c)$, if the computed anchor is a set of
features $\fml{A}$, then we should be able to validate that,
\begin{equation} \label{eq:chkaxp}
  \forall(\mbf{x}\in\mbb{F}).\bigwedge_{i\in\fml{A}}(x_i=v_i)\limply\kappa(\mbf{x})=c)
\end{equation}
which should be true if $\fml{A}$ were an AXp.
To prove that a set of features is not an AXp, we decide instead
whether the following stament is true,
\begin{equation} \label{eq:chkaxp2}
  \exists(\mbf{x}\in\mbb{F}).\bigwedge_{i\in\fml{A}}(x_i=v_i)\land\kappa(\mbf{x})\not=c)
\end{equation}
which can be formulated as the following decision problem,
\begin{equation} \label{eq:chkaxp3}
  \consistent{\bigwedge_{i\in\fml{A}}(x_i=v_i)\land\kappa(\mbf{x})\not=c)}
\end{equation}
where $\consistent{\varphi}$ decides the satisfiability of the logic
formula $\varphi$.
Depending on the logic representation of the classifier, one can use a
suitable decision procedure for deciding whether~\eqref{eq:chkaxp3}.
So, in practice, and given a set of features reported by
Anchor~\cite{guestrin-aaai18}, it suffices to use a decision procedure
to check whether~\eqref{eq:chkaxp} holds.
It should be noted that an alternative to deciding
whether~\eqref{eq:chkaxp} is to assess how likely it is
for~\eqref{eq:chkaxp} not to hold. This involves (exact or
approximate) model counting, and was investigated
in~\cite{nsmims-sat19}.

In the case of boosted trees~\cite{guestrin-kdd16b}, the
model-agnostic tool Anchor~\cite{guestrin-aaai18} was assessed, and
its explanations validated with a logic-based
explainer~\cite{inms-corr19,ignatiev-ijcai20}. The correctness
results are summarized in~\cref{tab:anchor} (more comprehensive
results are reported in earlier
work~\cite{inms-corr19,ignatiev-ijcai20,yisnms-aaai23}).
\begin{table}[t]
  \begin{center}
    \begin{tabular}{cccc} \toprule
      Dataset & \% Incorrect & \% Redundant & \% Correct \\
      \toprule
      adult & 80.5\% & 1.6\% & 17.9\% \\
      lending & 3.0\% & 0.0\% & 97.0\% \\
      rcdv & 99.4\% & 0.4\% & 0.2\% \\
      compas & 84.4\% & 1.7\% & 13.9\% \\
      german & 99.7\% & 0.2\% & 0.1\% \\
      \bottomrule
    \end{tabular}
  \end{center}
  \caption{Correctness results for Anchor's
    explanations~\cite{inms-corr19,ignatiev-ijcai20}}
  \label{tab:anchor}
\end{table}
Similar results were obtained by interpreting
LIME~\cite{guestrin-kdd16a} and SHAP~\cite{lundberg-nips17} as feature
selection explainers~\cite{ignatiev-ijcai20}.
A possible criticism of the results in~\cref{tab:anchor} is that all
inputs are assumed possible, and this may not always be the case.
More recent work reports less negative results on average, when
inferred background knowledge is taken into
account~\cite{yisnms-aaai23}.
However, even when inferred background knowledge is taken into
account, (non-formal) explainability with feature selection appears
unlikely to contribute to delivering trust in AI.
This remark is confirmed indirectly by results in~\cite{nsmims-sat19},
where the likelihood of picking the wrong prediction for computed
explanations was shown to be significant.

\jnoteF{Describe what is model-agnostic XAI.}

\jnoteF{Describe the Anchor approach.}

\jnoteF{Plan:
  \begin{enumerate}[nosep]
  \item The success of model agnostic methods
  \item A first question: are model-agnostic XPs sufficient for the
    prediction?
  \item What we have proved: most often this is not the case
  \item How did we do it?
  \item \tbf{Obs}: cite CoRR'19, IJCAI'20 \& AAAI'23 papers.
  \end{enumerate}
}

\section{Myth \#03: Shapley Values in Explainability}
\label{sec:myth03}

Shapley values have been studied as a method of feature attribution
for more than a decade~\cite{kononenko-jmlr10,kononenko-kis14},
attaining widespread visibility as the result of their use in
SHAP~\cite{lundberg-nips17,lundberg-naturemi20}.

Nevertheless, SHAP exploits approximate Shapley
values~\cite{lundberg-nips17}, given that the exact computation of
Shapley values is generally considered to be intractable.
Thus, and to the best of our knowledge, there is no validation of the
adequacy of Shapley values for explainability.
Moreover, recent work proved that Shapley values could be computed in
polynomial-time for restricted classes of boolean
circuits~\cite{barcelo-aaai21}.
In addition, when functions are represented by truth tables, then it
is simple to prove that Shapley values can be computed in polynomial
time on the size of the truth table~\cite{hms-corr23}.

Furthermore, if Shapley values correctly capture the relative
importance of features, then there should be a tight connection
between the values obtained by methods of feature attribution and the
features obtained with feature selection. Evidently, if a feature
\emph{cannot} be picked by a feature selection method, then it should
have \emph{no} assigned feature importance. 
Since most methods of feature selection are heuristic, offering no
guarantees of correctness, we consider formally defined abductive 
explanations as described in~\cref{ssec:fxai}.
Concretely, we focus on the (crucial) distinction between relevant and
irrelevant features and their relative importance as computed by
Shapley values for explainability. Moreover, we investigate whether
some relevant feature(s) could be assigned less importance that some
other irrelevant feature(s). Among a number of possible issues, we
investigated the following issues:
\begin{enumerate}[nosep,topsep=3.0pt,itemsep=3.0pt,label=\textbf{I\arabic*.},ref=\small\textrm{I\arabic*},leftmargin=0.75cm]  
\item For a boolean classifier, with an instance $(\mbf{v},c)$, and
  feature $i$ such that, \label{en:i1}
  \[
  \msf{Irrelevant}(i)\land\left(\msf{Sv}(i)\not=0\right)
  \]
  Thus, an~\cref{en:i1} issue is such that the feature is irrelevant,
  but its Shapley value is non-zero.
\item For a boolean classifier, with an instance $(\mbf{v},c)$ and
  features $i_1$ and $i_2$ such that, \label{en:i2}
  \[
  \begin{array}{l}
    \msf{Irrelevant}(i_1)\land\msf{Relevant}(i_2)\land
    \left(|\msf{Sv}(i_1)|>|\msf{Sv}(i_2)|\right)
  \end{array}
  \]
  Thus, an~\cref{en:i2} issue is such that there is at least one
  irrelevant feature exhibiting a Shapley value larger (in absolute
  value) than the Shapley of a relevant feature.
\item For a boolean classifier, with instance $(\mbf{v},c)$,
  and feature $i$ such that, \label{en:i3}
  \[
  \msf{Relevant}(i)\land\left(\msf{Sv}(i)=0\right)
  \]
  Thus, an~\cref{en:i3} issue is such that the feature is relevant,
  but its Shapley value is zero.
\item For a boolean classifier, with instance $(\mbf{v},c)$, and
  features $i_1$ and $i_2$ such that, \label{en:i4}
  \begin{align}
    [&\msf{Irrelevant}(i_1)\land\left(\msf{Sv}(i_1)\not=0\right)]
  \land\nonumber\\
  [&\msf{Relevant}(i_2)\land\left(\msf{Sv}(i_2)=0\right)] \nonumber
  \end{align}
  Thus, an~\cref{en:i4} issue is such that there is at least one
  irrelevant feature with a non-zero Shapley value and a relevant
  feature with a Shapley value of 0.
\item For a boolean classifier, with instance $(\mbf{v},c)$ and
  feature $i$ such that, \label{en:i5}
  \[
    [\msf{Irrelevant}(i)\land
      \forall_{1\le{j}\le{m},j\not=i}\left(|\msf{Sv}(j)|<|\msf{Sv}(i)|\right)]
  \]
  Thus, an~\cref{en:i5} issue is such that there is one irrelevant
  feature exhibiting the value Shapley value (in absolute value).
  (\cref{en:i5} can be viewed as a special case of the other issues,
  and is not analyzed separately in earlier work~\cite{hms-corr23}.)
\end{enumerate}
\begin{table*}[t]
  \begin{center}
    \scalebox{0.95}{\renewcommand{\tabcolsep}{0.5em}
\begin{tabular}{ccccC{2.75cm}c} \toprule
  Case & Instance & Relevant & Irrelevant & Shapley values & Justification
  \\ \toprule
  %
  %
  \cref{en:i4} &
  $((0,0,1,1),0)$ &
  $1,2,4$ &
  $3$ &
  $\begin{array}{l}\sv(1)=-0.13\\\sv(2)=0.33\\\sv(3)=0.08\\\sv(4)=0.00\\\end{array}$ &
  $\begin{array}{l}\irrelevant(3)\land\sv(3)\not=0\land\\\relevant(4)\land\sv(4)=0\end{array}$
  \\ \midrule
  \cref{en:i5} &
  $((1,1,1,1),0)$ &
  $1,2,3$ &
  $4$ &
  $\begin{array}{l}\sv(1)=-0.12\\\sv(2)=-0.12\\\sv(3)=-0.12\\\sv(4)=0.17\\\end{array}$ &
  $\begin{array}{l}\irrelevant(4)\land\\\forall(j\in\{1,2,3\}).|\sv(j)|<\sv(4)|\end{array}$
  \\ 
  
  \bottomrule
\end{tabular}
}
  \end{center}
  \caption{Examples of issues with Shapley for explainability for
    boolean classifiers of~\cref{fig:bfunc}} \label{tab:sv-issues}
\end{table*}
Whereas a more comprehensive assessment is available
elsewhere~\cite{hms-corr23}, the next example illustrates the
existence of issues~\cref{en:i4,en:i5} for the boolean classifiers
in~\cref{fig:bfunc}\footnote{%
Given a boolean classifier represented by a truth table,
\cite{hms-corr23} describes simple polynomial-time algorithms for
computing of Shapley values, one or more AXp's, one or more CXp's, and
for deciding relevancy/irrelevancy of features. The fact that the
classifier is represented by a truth table enables polynomial-time
algorithms for analyzing all subsets, computing average values, but
also deciding over existential/universal quantifiers}.

\begin{example}
  We focus on~\cref{en:i4,en:i5} (a more detailed analysis can be
  found elsewhere~\cite{hms-corr23}).
  \cref{tab:sv-issues} confirms the occurrence of issue~\cref{en:i4}
  for $\kappa_{I4}$ (see~\cref{ex:k4}), and of issue~\cref{en:i5} for
  $\kappa_{I5}$ (see~\cref{ex:k5}).
  Among these two cases, the example of~\cref{ex:k5} is significant,
  in that the irrelevant feature has the largest absolute Shapley
  value among all the features.
\end{example}

The example above (and the data reported in earlier
work~\cite{hms-corr23}) offer conclusive evidence that Shapley values
for explainability can provide misleading information, about the
relative importance of features, to human decision makers.


\jnoteF{Theoretical underpinnings of feature attribution based on
  Shapley values, since 2010, with great visibility since 2017... Lots
  of practical applications.}

\jnoteF{How to measure whether SVs can mislead decision makers?\\
  Our approach: look at feature relevancy, and find functions and
  instances where relevant features are less important than irrelevant
  features. Look for worrisome scenarios.
}

\jnoteF{Plan:
  \begin{enumerate}[nosep]
  \item The success of Sv's as the theoretical justification for SHAP
    and related tools.
  \item A first question: how to related feature attribution with
    feature selection.
  \item There is a well-defined characterization when referring to
    (rigorous) feature selection methods: features in some explanation
    and features excluded from all explanations, i.e.\ features that
    are relevant or irrelevant.
  \item A second question: do Sv's respect such characterization in
    terms of the relative order of feature importance?
  \item What we have proved: Sv's do not respect the distinction
    between relevant and irrelevant features.
  \item The issues \& the examples...
  \end{enumerate}
}

\section{Discussion} \label{sec:disc}

As noted earlier in~\cref{sec:myth01}, the refutation of myth \#01 is
based on the existence of redundant features in explanations produced
by manual inspection of DTs and DLs (a similar situation could be
envisioned for DSs).
One could argue that redundancy is not problematic per se, e.g.\ if
DTs are shallow or small. However, if redundancy of explanations were
acceptable, then reporting all features as an explanation could also
be argued to be acceptable. The key observation is that succinctness
and irreducibility are two desirable properties when reporting
explanations. Furthermore, there are recent examples of large deep DTs
in practical use~\cite{ghiasi-cmpb20}.

Regarding the other two myths, the situation is more problematic, in
that human decision makers can obtain erroneous information when using
either model-agnostic explainability (in the case of myth \#02) or
when using feature attribution based on Shapley values (in the case of
myth \#03).
If one critical goal of XAI is to build trust, then neither
model-agnostic explainability nor feature attribution with Shapley
values should be employed, in domains where trust matters, especially
in those where situations of catastrophic failure can occur.

\section{Conclusions \& Research Directions}
\label{sec:conc}

This paper overviews a number of recent results which disprove several
myths of (non-formal) XAI. 
The first myth of XAI is that simple models do not require
explanations since the model (which is called interpretable) is itself
the explanation~\cite{rudin-naturemi19,molnar-bk20,rudin-naturermp22}.
Recent results~\cite{iims-jair22,iims-corr20} prove in theory and
demonstrate in practice that redundant literals in paths are
ubiquitous in decision trees, one of the hallmarks of model
interpretability.
A second myth of XAI is the use of model-agnostic methods.
%
Recent work~\cite{inms-corr19,ignatiev-ijcai20,yisnms-aaai23} offers
experimental evidence that model-agnostic explanations are often
incorrect.
A third and final myth of XAI is the use of Shapley values as the
theoretical underpinning of (some) feature attribution methods.
%
Recent work~\cite{hms-corr23} demonstrated that Shapley values can
give misleading information regarding the relative importance of
features in explanations.

Future work in applying formal methods to XAI will continue to expand
the range of ML models that can be efficiently explained, but it will
continue to uncover other expected myths of XAI.

\acks*{} 
This work resulted from collaborations with several colleagues,
including
N.\ Asher,
M.\ Cooper, 
X.\ Huang,
A.\ Hurault,
A.\ Ignatiev,
Y.\ Izza,
O.\ L\'{e}toff\'{e},
C.\ Menc\'{\i}a,
A.\ Morgado,
N.\ Narodytska.
R.\ Passos
and
J.\ Planes.
%
%
The author also acknowledges
the incentive provided by the ERC who, by not funding this research
nor a handful of other grant applications between 2012 and 2022, has
had a lasting impact in framing the research presented in this paper.

%

\newtoggle{mkbbl}

\settoggle{mkbbl}{false}

\addcontentsline{toc}{section}{References}
\vskip 0.2in
\iftoggle{mkbbl}{
  \bibliographystyle{IEEEtranS}
  \bibliography{team,refs}

\begin{thebibliography}{10}
\providecommand{\url}[1]{#1}
\csname url@samestyle\endcsname
\providecommand{\newblock}{\relax}
\providecommand{\bibinfo}[2]{#2}
\providecommand{\BIBentrySTDinterwordspacing}{\spaceskip=0pt\relax}
\providecommand{\BIBentryALTinterwordstretchfactor}{4}
\providecommand{\BIBentryALTinterwordspacing}{\spaceskip=\fontdimen2\font plus
\BIBentryALTinterwordstretchfactor\fontdimen3\font minus
  \fontdimen4\font\relax}
\providecommand{\BIBforeignlanguage}[2]{{%
\expandafter\ifx\csname l@#1\endcsname\relax
\typeout{** WARNING: IEEEtranS.bst: No hyphenation pattern has been}%
\typeout{** loaded for the language `#1'. Using the pattern for}%
\typeout{** the default language instead.}%
\else
\language=\csname l@#1\endcsname
\fi
#2}}
\providecommand{\BIBdecl}{\relax}
\BIBdecl

\bibitem{amgoud-ijcai22}
L.~Amgoud and J.~Ben{-}Naim, ``Axiomatic foundations of explainability,'' in
  \emph{IJCAI}, 2022, pp. 636--642.

\bibitem{barcelo-nips21}
M.~Arenas, D.~Baez, P.~Barcel{\'{o}}, J.~P{\'{e}}rez, and B.~Subercaseaux,
  ``Foundations of symbolic languages for model interpretability,'' in
  \emph{NeurIPS}, 2021.

\bibitem{barcelo-corr21}
\BIBentryALTinterwordspacing
M.~Arenas, P.~Barcel{\'{o}}, L.~E. Bertossi, and M.~Monet, ``On the complexity
  of {SHAP}-score-based explanations: Tractability via knowledge compilation
  and non-approximability results,'' \emph{CoRR}, vol. abs/2104.08015, 2021.
  [Online]. Available: \url{https://arxiv.org/abs/2104.08015}
\BIBentrySTDinterwordspacing

\bibitem{barcelo-aaai21}
------, ``The tractability of {SHAP}-score-based explanations for
  classification over deterministic and decomposable boolean circuits,'' in
  \emph{AAAI}, 2021, pp. 6670--6678.

\bibitem{barcelo-corr22}
\BIBentryALTinterwordspacing
M.~Arenas, P.~Barcel{\'{o}}, M.~Romero, and B.~Subercaseaux, ``On computing
  probabilistic explanations for decision trees,'' \emph{CoRR}, vol.
  abs/2207.12213, 2022. [Online]. Available:
  \url{https://doi.org/10.48550/arXiv.2207.12213}
\BIBentrySTDinterwordspacing

\bibitem{marquis-kr21}
G.~Audemard, S.~Bellart, L.~Bounia, F.~Koriche, J.~Lagniez, and P.~Marquis,
  ``On the computational intelligibility of boolean classifiers,'' in
  \emph{KR}, 2021, pp. 74--86.

\bibitem{marquis-dke22}
------, ``On the explanatory power of boolean decision trees,'' \emph{Data
  Knowl. Eng.}, vol. 142, p. 102088, 2022.

\bibitem{muller-plosone15}
S.~Bach, A.~Binder, G.~Montavon, F.~Klauschen, K.-R. M{\"u}ller, and W.~Samek,
  ``On pixel-wise explanations for non-linear classifier decisions by
  layer-wise relevance propagation,'' \emph{PloS one}, vol.~10, no.~7, p.
  e0130140, 2015.

\bibitem{barcelo-nips20}
P.~Barcel{\'{o}}, M.~Monet, J.~P{\'{e}}rez, and B.~Subercaseaux, ``Model
  interpretability through the lens of computational complexity,'' in
  \emph{NeurIPS}, 2020.

\bibitem{fair-bk22}
S.~Barocas, M.~Hardt, and A.~Narayanan, \emph{Fairness in machine
  learning}.\hskip 1em plus 0.5em minus 0.4em\relax Online book, 2022,
  \url{https://fairmlbook.org/pdf/fairmlbook.pdf}.

\bibitem{bengio-cacm21}
\BIBentryALTinterwordspacing
Y.~Bengio, Y.~LeCun, and G.~E. Hinton, ``Deep learning for {AI},''
  \emph{Commun. {ACM}}, vol.~64, no.~7, pp. 58--65, 2021. [Online]. Available:
  \url{https://doi.org/10.1145/3448250}
\BIBentrySTDinterwordspacing

\bibitem{tan-nips21}
G.~Blanc, J.~Lange, and L.~Tan, ``Provably efficient, succinct, and precise
  explanations,'' in \emph{NeurIPS}, 2021.

\bibitem{mazure-cikm21}
R.~Boumazouza, F.~C. Alili, B.~Mazure, and K.~Tabia, ``{ASTERYX:} {A}
  model-agnostic {SaT}-based app{R}oach for s{Y}mbolic and score-based
  e{X}planations,'' in \emph{CIKM}, 2021, pp. 120--129.

\bibitem{breiman-ss01}
L.~Breiman, ``Statistical modeling: The two cultures,'' \emph{Statistical
  science}, vol.~16, no.~3, pp. 199--231, 2001.

\bibitem{jordan-iclr19}
J.~Chen, L.~Song, M.~J. Wainwright, and M.~I. Jordan, ``{L}-{S}hapley and
  {C}-{S}hapley: Efficient model interpretation for structured data,'' in
  \emph{ICLR}, 2019.

\bibitem{guestrin-kdd16b}
T.~Chen and C.~Guestrin, ``{XGBoost}: {A} scalable tree boosting system,'' in
  \emph{KDD}, 2016, pp. 785--794.

\bibitem{cms-cp21}
M.~C. Cooper and J.~Marques{-}Silva, ``On the tractability of explaining
  decisions of classifiers,'' in \emph{CP}, L.~D. Michel, Ed., 2021, pp.
  21:1--21:18.

\bibitem{cms-aij23}
\BIBentryALTinterwordspacing
------, ``Tractability of explaining classifier decisions,'' \emph{Artif.
  Intell.}, vol. 316, p. 103841, 2023. [Online]. Available:
  \url{https://doi.org/10.1016/j.artint.2022.103841}
\BIBentrySTDinterwordspacing

\bibitem{darwiche-jlli23}
\BIBentryALTinterwordspacing
A.~Darwiche and A.~Hirth, ``On the (complete) reasons behind decisions,''
  \emph{J. Log. Lang. Inf.}, vol.~32, no.~1, pp. 63--88, 2023. [Online].
  Available: \url{https://doi.org/10.1007/s10849-022-09377-8}
\BIBentrySTDinterwordspacing

\bibitem{darwiche-jair21}
A.~Darwiche and P.~Marquis, ``On quantifying literals in boolean logic and its
  applications to explainable {AI},'' \emph{J. Artif. Intell. Res.}, vol.~72,
  pp. 285--328, 2021.

\bibitem{zick-sp16}
A.~Datta, S.~Sen, and Y.~Zick, ``Algorithmic transparency via quantitative
  input influence: Theory and experiments with learning systems,'' in
  \emph{IEEE S\&P}, 2016, pp. 598--617.

\bibitem{vandenbroeck-aaai21}
G.~V. den Broeck, A.~Lykov, M.~Schleich, and D.~Suciu, ``On the tractability of
  {SHAP} explanations,'' in \emph{AAAI}, 2021, pp. 6505--6513.

\bibitem{vandenbroeck-jair22}
\BIBentryALTinterwordspacing
------, ``On the tractability of {SHAP} explanations,'' \emph{J. Artif. Intell.
  Res.}, vol.~74, pp. 851--886, 2022. [Online]. Available:
  \url{https://doi.org/10.1613/jair.1.13283}
\BIBentrySTDinterwordspacing

\bibitem{leite-kr22}
J.~Ferreira, M.~de~Sousa~Ribeiro, R.~Gon{\c{c}}alves, and J.~Leite, ``Looking
  inside the black-box: Logic-based explanations for neural networks,'' in
  \emph{KR}, 2022, p. 432–442.

\bibitem{ghiasi-cmpb20}
\BIBentryALTinterwordspacing
M.~M. Ghiasi, S.~Zendehboudi, and A.~A. Mohsenipour, ``Decision tree-based
  diagnosis of coronary artery disease: {CART} model,'' \emph{Comput. Methods
  Programs Biomed.}, vol. 192, p. 105400, 2020. [Online]. Available:
  \url{https://doi.org/10.1016/j.cmpb.2020.105400}
\BIBentrySTDinterwordspacing

\bibitem{rubin-aaai22}
N.~Gorji and S.~Rubin, ``Sufficient reasons for classifier decisions in the
  presence of domain constraints,'' in \emph{{AAAI}}, February 2022.

\bibitem{pedreschi-acmcs19}
R.~Guidotti, A.~Monreale, S.~Ruggieri, F.~Turini, F.~Giannotti, and
  D.~Pedreschi, ``A survey of methods for explaining black box models,''
  \emph{{ACM} Comput. Surv.}, vol.~51, no.~5, pp. 93:1--93:42, 2019.

\bibitem{gunning-tr16}
D.~Gunning, ``Explainable artificial intelligence {XAI()},'' DARPA, Tech. Rep.
  DARPA-BAA-16-53, 8 2016.

\bibitem{gunning-aimag19}
\BIBentryALTinterwordspacing
D.~Gunning and D.~W. Aha, ``{DARPA}'s explainable artificial intelligence
  {(XAI)} program,'' \emph{{AI} Mag.}, vol.~40, no.~2, pp. 44--58, 2019.
  [Online]. Available: \url{https://doi.org/10.1609/aimag.v40i2.2850}
\BIBentrySTDinterwordspacing

\bibitem{gunning-sr19}
\BIBentryALTinterwordspacing
D.~Gunning, M.~Stefik, J.~Choi, T.~Miller, S.~Stumpf, and G.~Yang, ``{XAI} -
  explainable artificial intelligence,'' \emph{Sci. Robotics}, vol.~4, no.~37,
  2019. [Online]. Available: \url{https://doi.org/10.1126/scirobotics.aay7120}
\BIBentrySTDinterwordspacing

\bibitem{rudin-nips19}
X.~Hu, C.~Rudin, and M.~I. Seltzer, ``Optimal sparse decision trees,'' in
  \emph{NeurIPS}, 2019, pp. 7265--7273.

\bibitem{hcmpms-tacas23}
X.~Huang, M.~C. Cooper, A.~Morgado, J.~Planes, and J.~Marques{-}Silva,
  ``Feature necessity \& relevancy in {ML} classifier explanations,'' in
  \emph{TACAS}, 4 2023.

\bibitem{hiicams-aaai22}
X.~Huang, Y.~Izza, A.~Ignatiev, M.~C. Cooper, N.~Asher, and J.~Marques{-}Silva,
  ``Tractable explanations for {d-DNNF} classifiers,'' in \emph{AAAI}, 2022,
  pp. 5719--5728.

\bibitem{hiims-kr21}
X.~Huang, Y.~Izza, A.~Ignatiev, and J.~Marques{-}Silva, ``On efficiently
  explaining graph-based classifiers,'' in \emph{KR}, 2021, pp. 356--367.

\bibitem{hims-aaai23}
X.~Huang, Y.~Izza, and J.~Marques-Silva, ``Solving explainability queries with
  quantification: The case of feature relevancy,'' in \emph{{AAAI}}, 2023.

\bibitem{hms-corr23}
\BIBentryALTinterwordspacing
X.~Huang and J.~Marques{-}Silva, ``The inadequacy of {S}hapley values for
  explainability,'' \emph{CoRR}, vol. abs/2302.08160, 2023. [Online].
  Available: \url{https://doi.org/10.48550/arXiv.2302.08160}
\BIBentrySTDinterwordspacing

\bibitem{ignatiev-ijcai20}
A.~Ignatiev, ``Towards trustable explainable {AI},'' in \emph{IJCAI}, 2020, pp.
  5154--5158.

\bibitem{icshms-cp20}
A.~Ignatiev, M.~C. Cooper, M.~Siala, E.~Hebrard, and J.~Marques{-}Silva,
  ``Towards formal fairness in machine learning,'' in \emph{CP}, 2020, pp.
  846--867.

\bibitem{iisms-aaai22}
A.~Ignatiev, Y.~Izza, P.~J. Stuckey, and J.~Marques{-}Silva, ``Using {MaxSAT}
  for efficient explanations of tree ensembles,'' in \emph{AAAI}, 2022, pp.
  3776--3785.

\bibitem{ims-sat21}
A.~Ignatiev and J.~Marques{-}Silva, ``{SAT}-based rigorous explanations for
  decision lists,'' in \emph{SAT}, 2021, pp. 251--269.

\bibitem{inams-aiia20}
A.~Ignatiev, N.~Narodytska, N.~Asher, and J.~Marques{-}Silva, ``From
  contrastive to abductive explanations and back again,'' in \emph{AIxIA},
  2020, pp. 335--355.

\bibitem{inms-aaai19}
A.~Ignatiev, N.~Narodytska, and J.~Marques{-}Silva, ``Abduction-based
  explanations for machine learning models,'' in \emph{AAAI}, 2019, pp.
  1511--1519.

\bibitem{inms-nips19}
------, ``On relating explanations and adversarial examples,'' in
  \emph{NeurIPS}, 2019, pp. 15\,857--15\,867.

\bibitem{inms-corr19}
\BIBentryALTinterwordspacing
------, ``On validating, repairing and refining heuristic {ML} explanations,''
  \emph{CoRR}, vol. abs/1907.02509, 2019. [Online]. Available:
  \url{http://arxiv.org/abs/1907.02509}
\BIBentrySTDinterwordspacing

\bibitem{iims-corr20}
\BIBentryALTinterwordspacing
Y.~Izza, A.~Ignatiev, and J.~Marques{-}Silva, ``On explaining decision trees,''
  \emph{CoRR}, vol. abs/2010.11034, 2020. [Online]. Available:
  \url{https://arxiv.org/abs/2010.11034}
\BIBentrySTDinterwordspacing

\bibitem{iims-jair22}
\BIBentryALTinterwordspacing
------, ``On tackling explanation redundancy in decision trees,'' \emph{J.
  Artif. Intell. Res.}, vol.~75, pp. 261--321, 2022. [Online]. Available:
  \url{https://jair.org/index.php/jair/article/view/13575/}
\BIBentrySTDinterwordspacing

\bibitem{iims-corr22}
\BIBentryALTinterwordspacing
------, ``On tackling explanation redundancy in decision trees,'' \emph{CoRR},
  vol. abs/2205.09971, 2022. [Online]. Available:
  \url{https://doi.org/10.48550/arXiv.2205.09971}
\BIBentrySTDinterwordspacing

\bibitem{ims-ijcai21}
Y.~Izza and J.~Marques{-}Silva, ``On explaining random forests with {SAT},'' in
  \emph{IJCAI}, 2021, pp. 2584--2591.

\bibitem{leskovec-kdd16}
H.~Lakkaraju, S.~H. Bach, and J.~Leskovec, ``Interpretable decision sets: {A}
  joint framework for description and prediction,'' in \emph{KDD}, 2016, pp.
  1675--1684.

\bibitem{bengio-nature15}
Y.~LeCun, Y.~Bengio, and G.~Hinton, ``Deep learning,'' \emph{nature}, vol. 521,
  no. 7553, pp. 436--444, 2015.

\bibitem{lipton-cacm18}
Z.~C. Lipton, ``The mythos of model interpretability,'' \emph{Commun. {ACM}},
  vol.~61, no.~10, pp. 36--43, 2018.

\bibitem{lorini-jlc23}
\BIBentryALTinterwordspacing
X.~Liu and E.~Lorini, ``A unified logical framework for explanations in
  classifier systems,'' \emph{J. Log. Comput.}, vol.~33, no.~2, pp. 485--515,
  2023. [Online]. Available: \url{https://doi.org/10.1093/logcom/exac102}
\BIBentrySTDinterwordspacing

\bibitem{lundberg-naturemi20}
\BIBentryALTinterwordspacing
S.~M. Lundberg, G.~G. Erion, H.~Chen, A.~J. DeGrave, J.~M. Prutkin, B.~Nair,
  R.~Katz, J.~Himmelfarb, N.~Bansal, and S.~Lee, ``From local explanations to
  global understanding with explainable {AI} for trees,'' \emph{Nat. Mach.
  Intell.}, vol.~2, no.~1, pp. 56--67, 2020. [Online]. Available:
  \url{https://doi.org/10.1038/s42256-019-0138-9}
\BIBentrySTDinterwordspacing

\bibitem{lundberg-nips17}
S.~M. Lundberg and S.~Lee, ``A unified approach to interpreting model
  predictions,'' in \emph{NeurIPS}, 2017, pp. 4765--4774.

\bibitem{kwiatkowska-ijcai21}
E.~L. Malfa, R.~Michelmore, A.~M. Zbrzezny, N.~Paoletti, and M.~Kwiatkowska,
  ``On guaranteed optimal robust explanations for {NLP} models,'' in
  \emph{IJCAI}, 2021, pp. 2658--2665.

\bibitem{ms-corr22}
\BIBentryALTinterwordspacing
J.~Marques{-}Silva, ``Logic-based explainability in machine learning,''
  \emph{CoRR}, vol. abs/2211.00541, 2022. [Online]. Available:
  \url{https://doi.org/10.48550/arXiv.2211.00541}
\BIBentrySTDinterwordspacing

\bibitem{msgcin-nips20}
J.~Marques{-}Silva, T.~Gerspacher, M.~C. Cooper, A.~Ignatiev, and
  N.~Narodytska, ``Explaining naive bayes and other linear classifiers with
  polynomial time and delay,'' in \emph{NeurIPS}, 2020.

\bibitem{msgcin-icml21}
------, ``Explanations for monotonic classifiers,'' in \emph{ICML}, 2021, pp.
  7469--7479.

\bibitem{msi-aaai22}
J.~Marques{-}Silva and A.~Ignatiev, ``Delivering trustworthy {AI} through
  formal {XAI},'' in \emph{AAAI}, 2022, pp. 12\,342--12\,350.

\bibitem{msi-fai23}
\BIBentryALTinterwordspacing
J.~Marques-Silva and A.~Ignatiev, ``No silver bullet: interpretable ml models
  must be explained,'' \emph{Frontiers in Artificial Intelligence}, vol.~6,
  2023. [Online]. Available:
  \url{https://www.frontiersin.org/articles/10.3389/frai.2023.1128212}
\BIBentrySTDinterwordspacing

\bibitem{galstyan-acmcs21}
\BIBentryALTinterwordspacing
N.~Mehrabi, F.~Morstatter, N.~Saxena, K.~Lerman, and A.~Galstyan, ``A survey on
  bias and fairness in machine learning,'' \emph{{ACM} Comput. Surv.}, vol.~54,
  no.~6, pp. 115:1--115:35, 2021. [Online]. Available:
  \url{https://doi.org/10.1145/3457607}
\BIBentrySTDinterwordspacing

\bibitem{taly-cdmake20}
L.~Merrick and A.~Taly, ``The explanation game: Explaining machine learning
  models using {S}hapley values,'' in \emph{CDMAKE}, 2020, pp. 17--38.

\bibitem{miller-pr56}
G.~A. Miller, ``The magical number seven, plus or minus two: Some limits on our
  capacity for processing information.'' \emph{Psychological review}, vol.~63,
  no.~2, pp. 81--97, 1956.

\bibitem{miller-aij19}
T.~Miller, ``Explanation in artificial intelligence: Insights from the social
  sciences,'' \emph{Artif. Intell.}, vol. 267, pp. 1--38, 2019.

\bibitem{molnar-bk20}
C.~Molnar, \emph{Interpretable Machine Learning}.\hskip 1em plus 0.5em minus
  0.4em\relax Leanpub, 2020, \url{http://tiny.cc/6c76tz}.

\bibitem{muller-dsp18}
G.~Montavon, W.~Samek, and K.~M{\"{u}}ller, ``Methods for interpreting and
  understanding deep neural networks,'' \emph{Digit. Signal Process.}, vol.~73,
  pp. 1--15, 2018.

\bibitem{nsmims-sat19}
N.~Narodytska, A.~A. Shrotri, K.~S. Meel, A.~Ignatiev, and J.~Marques{-}Silva,
  ``Assessing heuristic machine learning explanations with model counting,'' in
  \emph{SAT}, 2019, pp. 267--278.

\bibitem{shmueli-acmcs22}
\BIBentryALTinterwordspacing
D.~Pessach and E.~Shmueli, ``A review on fairness in machine learning,''
  \emph{{ACM} Comput. Surv.}, vol.~55, no.~3, pp. 51:1--51:44, 2023. [Online].
  Available: \url{https://doi.org/10.1145/3494672}
\BIBentrySTDinterwordspacing

\bibitem{propublica16}
ProPublica, ``Machine bias,'' \url{https://bit.ly/3zaHb2F}, 2016.

\bibitem{reiter-aij87}
\BIBentryALTinterwordspacing
R.~Reiter, ``A theory of diagnosis from first principles,'' \emph{Artif.
  Intell.}, vol.~32, no.~1, pp. 57--95, 1987. [Online]. Available:
  \url{https://doi.org/10.1016/0004-3702(87)90062-2}
\BIBentrySTDinterwordspacing

\bibitem{guestrin-kdd16a}
M.~T. Ribeiro, S.~Singh, and C.~Guestrin, ``"why should {I} trust you?":
  Explaining the predictions of any classifier,'' in \emph{KDD}, 2016, pp.
  1135--1144.

\bibitem{guestrin-aaai18}
------, ``Anchors: High-precision model-agnostic explanations,'' in
  \emph{AAAI}, 2018, pp. 1527--1535.

\bibitem{rudin-naturemi19}
C.~Rudin, ``Stop explaining black box machine learning models for high stakes
  decisions and use interpretable models instead,'' \emph{Nature Machine
  Intelligence}, vol.~1, no.~5, pp. 206--215, 2019.

\bibitem{rudin-naturermp22}
------, ``Why black box machine learning should be avoided for high-stakes
  decisions, in brief,'' \emph{Nature Reviews Methods Primers}, vol.~2, no.~1,
  pp. 1--2, 2022.

\bibitem{rudin-ss22}
C.~Rudin, C.~Chen, Z.~Chen, H.~Huang, L.~Semenova, and C.~Zhong,
  ``Interpretable machine learning: Fundamental principles and 10 grand
  challenges,'' \emph{Statistics Surveys}, vol.~16, pp. 1--85, 2022.

\bibitem{muller-ieee-proc21}
\BIBentryALTinterwordspacing
W.~Samek, G.~Montavon, S.~Lapuschkin, C.~J. Anders, and K.~M{\"{u}}ller,
  ``Explaining deep neural networks and beyond: {A} review of methods and
  applications,'' \emph{Proc. {IEEE}}, vol. 109, no.~3, pp. 247--278, 2021.
  [Online]. Available: \url{https://doi.org/10.1109/JPROC.2021.3060483}
\BIBentrySTDinterwordspacing

\bibitem{xai-bk19}
W.~Samek, G.~Montavon, A.~Vedaldi, L.~K. Hansen, and K.~M{\"{u}}ller, Eds.,
  \emph{Explainable {AI:} Interpreting, Explaining and Visualizing Deep
  Learning}.\hskip 1em plus 0.5em minus 0.4em\relax Springer, 2019.

\bibitem{muller-xai19-ch01}
W.~Samek and K.~M{\"{u}}ller, ``Towards explainable artificial intelligence,''
  in \emph{Explainable {AI:} Interpreting, Explaining and Visualizing Deep
  Learning}, W.~Samek, G.~Montavon, A.~Vedaldi, L.~K. Hansen, and
  K.~M{\"{u}}ller, Eds.\hskip 1em plus 0.5em minus 0.4em\relax Springer, 2019,
  pp. 5--22.

\bibitem{seshia-cacm22}
\BIBentryALTinterwordspacing
S.~A. Seshia, D.~Sadigh, and S.~S. Sastry, ``Toward verified artificial
  intelligence,'' \emph{Commun. {ACM}}, vol.~65, no.~7, pp. 46--55, 2022.
  [Online]. Available: \url{https://doi.org/10.1145/3503914}
\BIBentrySTDinterwordspacing

\bibitem{shapley-ctg53}
L.~S. Shapley, ``A value for $n$-person games,'' \emph{Contributions to the
  Theory of Games}, vol.~2, no.~28, pp. 307--317, 1953.

\bibitem{darwiche-ijcai18}
A.~Shih, A.~Choi, and A.~Darwiche, ``A symbolic approach to explaining bayesian
  network classifiers,'' in \emph{IJCAI}, 2018, pp. 5103--5111.

\bibitem{lakkaraju-nips21}
D.~Slack, A.~Hilgard, S.~Singh, and H.~Lakkaraju, ``Reliable post hoc
  explanations: Modeling uncertainty in explainability,'' in \emph{NeurIPS},
  2021, pp. 9391--9404.

\bibitem{kononenko-jmlr10}
\BIBentryALTinterwordspacing
E.~Strumbelj and I.~Kononenko, ``An efficient explanation of individual
  classifications using game theory,'' \emph{J. Mach. Learn. Res.}, vol.~11,
  pp. 1--18, 2010. [Online]. Available:
  \url{https://dl.acm.org/doi/10.5555/1756006.1756007}
\BIBentrySTDinterwordspacing

\bibitem{kononenko-kis14}
\BIBentryALTinterwordspacing
------, ``Explaining prediction models and individual predictions with feature
  contributions,'' \emph{Knowl. Inf. Syst.}, vol.~41, no.~3, pp. 647--665,
  2014. [Online]. Available: \url{https://doi.org/10.1007/s10115-013-0679-x}
\BIBentrySTDinterwordspacing

\bibitem{goodfellow-iclr14}
C.~Szegedy, W.~Zaremba, I.~Sutskever, J.~Bruna, D.~Erhan, I.~J. Goodfellow, and
  R.~Fergus, ``Intriguing properties of neural networks,'' in \emph{ICLR},
  2014.

\bibitem{theverge20}
{The Verge}, ``{UK} ditches exam results generated by biased algorithm after
  student protests,'' \url{https://bit.ly/3kevIsB}, 2020.

\bibitem{kutyniok-jair21}
\BIBentryALTinterwordspacing
S.~W{\"{a}}ldchen, J.~MacDonald, S.~Hauch, and G.~Kutyniok, ``The computational
  complexity of understanding binary classifier decisions,'' \emph{J. Artif.
  Intell. Res.}, vol.~70, pp. 351--387, 2021. [Online]. Available:
  \url{https://doi.org/10.1613/jair.1.12359}
\BIBentrySTDinterwordspacing

\bibitem{watson-facct22}
D.~S. Watson, ``Rational {S}hapley values,'' in \emph{FAccT}, 2022, pp.
  1083--1094.

\bibitem{yisnms-aaai23}
J.~Yu, A.~Ignatiev, P.~J. Stuckey, N.~Narodytska, and J.~Marques-Silva,
  ``Eliminating the impossible, whatever remains must be true: On extracting
  and applying background knowledge in the context of formal explanations,'' in
  \emph{{AAAI}}, 2023.

\end{thebibliography}
}{

}
\label{lastpage}

\end{document}